\PassOptionsToPackage{dvipsnames,table}{xcolor}
\documentclass{article}
\usepackage{log_2024}						

\usepackage{float}
\usepackage{amsfonts}
\usepackage{amssymb}
\usepackage{amsthm}
\usepackage{nicefrac}
\usepackage{microtype}  
\usepackage{multirow}
\usepackage{makecell}
\usepackage{mathtools}

\usepackage{bm}
\usepackage[linesnumbered,ruled,vlined]{algorithm2e}
\usepackage{siunitx}
\usepackage{booktabs}
\usepackage{array}
\newcolumntype{P}[1]{>{\centering\arraybackslash}p{#1}}
\SetCommentSty{mycommfont}
\DeclareMathOperator*{\argmin}{arg\,min}
\SetKwInput{KwInput}{Input}                
\SetKwInput{KwOutput}{Output}              

\usepackage{subfigure}
\usepackage{subcaption}
\usepackage{url}

\usepackage{natbib}
\theoremstyle{plain}
\newtheorem{theorem}{Theorem}[section]

\theoremstyle{definition}
\newtheorem{definition}[theorem]{Definition}

\theoremstyle{remark}
\newtheorem{remark}[theorem]{Remark}

\usepackage[textsize=tiny]{todonotes}
\title{T-GAE: Transferable Graph Autoencoder for Network Alignment}

\author[J. He et al.]{%
Jiashu He\\
University of Pennsylvania\\
\email{jiashuhe@seas.upenn.edu}\And
Charilaos Kanatsoulis\\
Stanford University\\
\email{charilaos@cs.stanford.edu}\And
Alejandro Ribeiro\\
University of Pennsylvania\\
\email{aribeiro@seas.upenn.edu}
}

\begin{document}

\maketitle

\begin{abstract}
 Network alignment is the task of establishing one-to-one correspondences between the nodes of different graphs. Although finding a plethora of applications in high-impact domains, this task is known to be NP-hard in its general form. Existing optimization algorithms do not scale up as the size of the graphs increases. While being able to reduce the matching complexity, current GNN approaches fit a deep neural network on each graph and requires re-train on unseen samples, which is time and memory inefficient. To tackle both challenges we propose T-GAE, a transferable graph autoencoder framework that leverages transferability and stability of GNNs to achieve efficient network alignment on out-of-distribution graphs without retraining. We prove that GNN-generated embeddings can achieve more accurate alignment compared to classical spectral methods. Our experiments on real-world benchmarks demonstrate that T-GAE outperforms the state-of-the-art optimization method and the best GNN approach by up to 38.7\% and 50.8\%, respectively, while being able to reduce 90\% of the training time when matching out-of-distribution large scale networks. We conduct ablation studies to highlight the effectiveness of the proposed encoder architecture and training objective in enhancing the expressiveness of GNNs to match perturbed graphs. T-GAE is also proved to be flexible to utilize matching algorithms of different complexities. Our code is available at \url{https://github.com/Jason-Tree/T-GAE}.
\end{abstract}

\section{Introduction}
Network alignment, also known as graph matching, is a classical problem in graph theory, that aims to find node correspondence across different graphs and is vital in a number of high-impact domains \citep{emmert2016fifty}. In social networks, for instance, network alignment has been used for user deanonymization \citep{nilizadeh2014community} and analysis \citep{Ogaard}, while in bioinformatics it is a key tool to identify functionalities in protein complexes \citep{bio2}, or to identify gene–drug modules \citep{10.1093/bioinformatics/bty662}. Graph matching also finds application in computer vision \citep{1246606}, sociology \citep{2021_baf4f1a5}, to name a few. However, this problem is usually cast as a quadratic assignment problem (QAP), which is in general NP-hard. 

Various approaches have been developed to tackle network alignment and can be divided into two main categories; i) optimization algorithms that attempt to approximate the QAP problem by relaxing the combinatorial constraints, ii) embedding methods that approach the problem by implicitly or explicitly generating powerful node embeddings that facilitate the alignment task. Optimization approaches, as \citep{anstreicher2001solving,vogelstein2015fast} employ quadratic programming relaxations, while  \citep{klau2009new} and \citep{peng2010new} utilize semidefinite or Lagrangian-based relaxations respectively, \citep{du1} and \citep{du2} proposed to solve network alignment together with link prediction. Successive convex approximations were also proposed by \citep{konar2020graph} to handle the QAP. Challenges associated with these methods include high computational cost, infeasible solutions, or nearly optimal initialization requirements. Embedding methods, on the other hand, overcome these challenges, but they usually produce inferior solutions, due to an inherent trade-off between embedding permutation-equivariance and the ability to capture the structural information of the graph. Typical embedding techniques include spectral and factorization methods \citep{6778, feizi2019spectral,onlineoffline,kanatsoulis2022gage}, structural feature engineering methods \citep{netsimile,heimann2018regal}, and random walk approaches \citep{deepwalk,grover2016node2vec}. Recently \citep{chen2020cone,karakasis2021joint} have proposed joint node embedding and network alignment, to overcome these challenges, but these methods do not scale up as the size of the graph increases.
\vspace{-0.1cm}
 \begin{figure*}
\centering
  \includegraphics[width=0.9\textwidth]{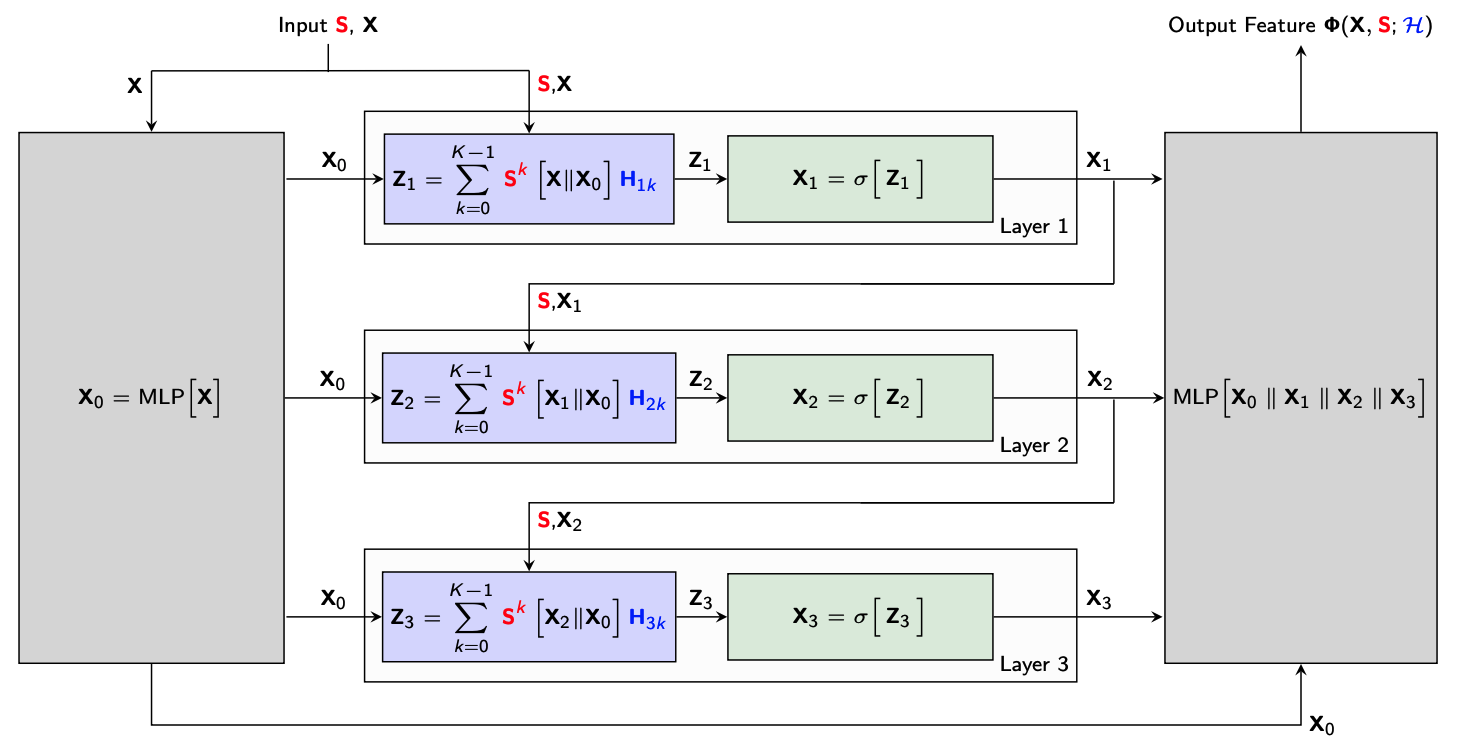}
  \caption{To enhance the expressiveness of GNNs to align the unseen graphs, our proposed encoder processes the input features by a local MLP, we then incorporate attention mechanism on (1) input to each message-passing layer by attending to the output of the last layer and processed input feature. (2) output of the encoder by attending to the output of each message passing layer.}
  \label{Model}
  \vspace{-0.4cm}
\end{figure*}

Graph Neural Networks (GNNs) are powerful architectures that learn graph representations (embeddings) in a self-supervised way \citep{GAE, GAE2}. They have shown state-of-the-art performance in several tasks, including biomedical \citep{Gainza2020, Strokach2020, Jiang2021, hu2023explainable, gtp4o}, quantum chemistry \citep{gilmer2017neural}, social networks and recommender systems \citep{Ying2018,survey, yang1, yang2}. A line of studies has been conducted to formulate key-point matching on images as graph matching problems \citep{keypoint1, keypoint2}. These frameworks rely on a powerful domain-specific encoder (CNNs, for example) to provide high-quality features. Given these high-quality features, GNNs are able to match graphs without training \citep{notraining}. However, expressive features are expensive and sometimes infeasible to build \citep{feature1, feature2}. \citep{cross} uses another trainable matrix to parameterize the internal connectivity between nodes of different graphs for faithful node alignment. Recently, \citep{jundong} proposed use GNNs to learn node embedding, and match nodes with small Wasserstein distances. However, this method needs to fit a GNN on every input graph, which results in very high training cost, since training deep GNNs with large sizes graphs is computationally prohibitive, as GNNs have limited scalability with respect to graph and model sizes \citep{GNNscale1, GNNscale2, GNNscale3, GNNscale4}.

To address these challenges, we propose T-GAE, a novel self-supervised GNN framework to perform network alignment. Specifically, we propose to utilize the transferability and robuseness of GNN to produce permutation equivariant and highly expressive embeddings. T-GAE trains the encoder on multiple families of small graphs and produce expressive/permutation equivariant representations for larger \textbf{\textit{unseen}} networks. We further prove that GNN representations combine the eigenvectors of the graph in a nonlinear fashion and are at least as good in network alignment as certain spectral methods. T-GAE is a one-shot solution that tackles the challenges of real-time network alignment from (1) Optimization based algorithms: high computational cost, assume ground truth node correspondence as initialization. (2) Deep-learning based frameworks: re-train for every pair of graphs, rely on high quality of node features. Extensive experiments with real-world benchmarks demonstrate the effectiveness and efficiency of the proposed approach in both graph and sub-graph matching, thereby sheds light on the potential of GNN to tackle the highly-complex network optimization problems.
\vspace{-0.2cm}
\section{Preliminaries}
Graphs are represented by $\mathcal{G}:=(\mathcal{V},\mathcal{E})$, where $\mathcal{V}=\{1,\dots,N\}$ is the set of vertices (nodes) and $\mathcal{E}=\left\{\left(v,u\right)\right\}$ correspond to edges between pairs of vertices. A graph is represented in a matrix form by a graph operator $\bm S\in\mathbb{R}^{N\times N}$, where $\bm S(i,j)$ quantifies the relation between node $i$ and node $j$ and $N = |\mathcal{V}|$ is the total number of vertices. In this work, we use the normalized graph adjacency and study the most general form of network alignment where there is no given graph attributes and ground truth node correspondence(anchor links).
\subsection{Network Alignment}
\begin{definition}[Network Alignment]
\label{def:GM}
Given a pair of graphs $\mathcal{G}:=(\mathcal{V},\mathcal{E}),~\hat{\mathcal{G}}:=(\hat{\mathcal{V}},\hat{\mathcal{E}})$, with graph adjacencies $\bm{S},~\hat{\bm{S}}$, network alignment aims to find a bijection $g: \mathcal{V}\rightarrow \hat{\mathcal{V}}$ which minimizes the number of edge disagreements between the two graphs. Formally, the problem can be written as:
\begin{equation}\label{eq:quadraticassignment}  \min_{\bm P\in\mathcal{P}}~\left\lVert~\bm S-\bm P\hat{\bm S}\bm P^T~\right\rVert_F^2,
\end{equation}
where $\mathcal{P}$ is the set of permutation matrices.
\end{definition}
As mentioned in the introduction, network alignment, is equivalent to the QAP, which has been proven to be NP-hard \citep{koopmans1957assignment}.
\begin{figure*}
  \includegraphics[width=\textwidth]{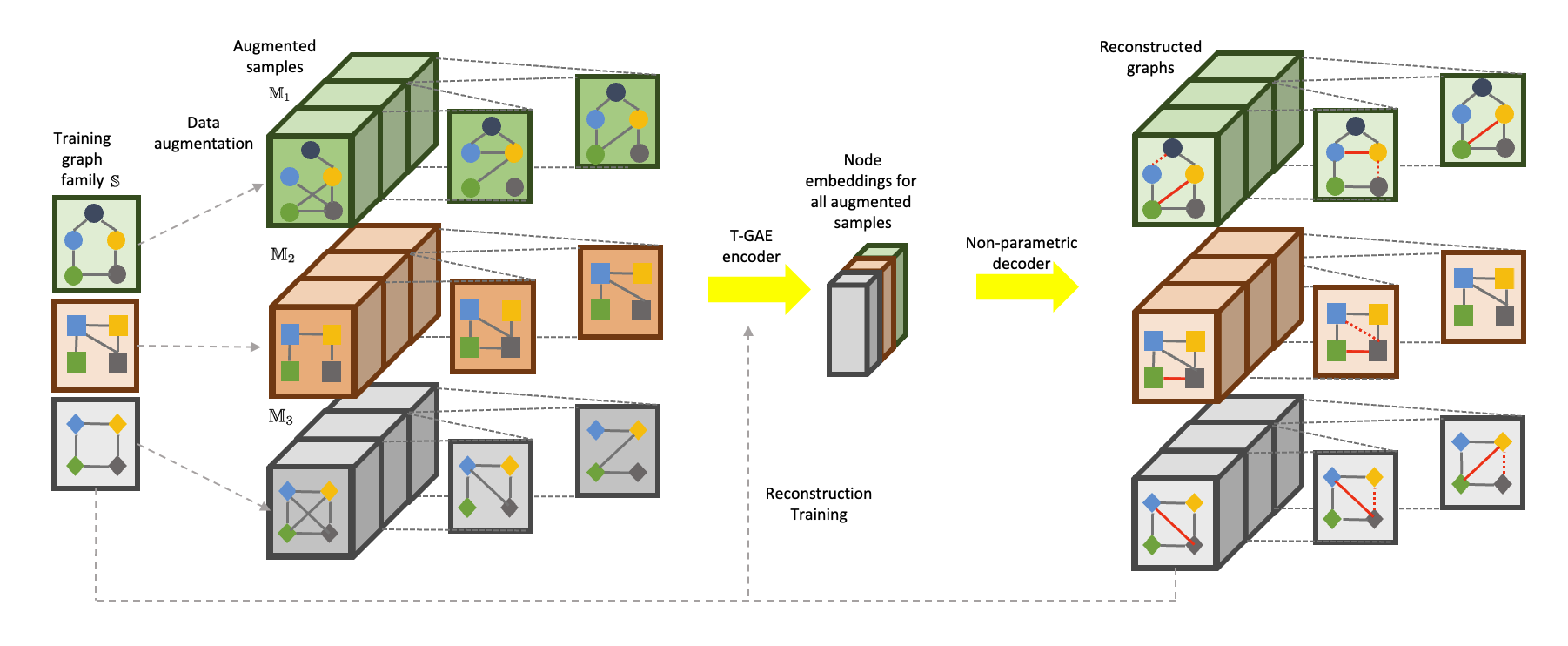}
  \vspace{-0.5cm}
  \caption{Proposed training objective: For each of the training graph, we generate a number of augmented samples by randomly adding or removing edges. The node embedding of these augmented samples are decoded non-parametrically, and compared with the corresponding original graph to train the T-GAE encoder.}
  \label{TrainingPic}
  \vspace{-0.6cm}
\end{figure*}

\subsection{Spectral Decomposition of the Graph}
A popular approach to tackle network alignment is by learning powerful node embeddings associated with connectivity information in the graph. Network alignment can be achieved by matching the node embeddings of different graphs rather than graph adjacencies, as follows:
\begin{equation}\label{eq:linearassignment} \min_{\bm P\in\mathcal{P}}~\left\lVert~\bm E-\bm P\hat{\bm E}~\right\rVert_F^2,
\end{equation}
where $\bm E\in\mathbb{R}^{N\times F}$ is embedding matrix and $\bm E[i,:]$ is the vector representation of node $i$. The optimization problem in \eqref{eq:linearassignment} is a linear assignment problem and can be optimally solved in $\mathcal{O}\left(N^3\right)$ by the Hungarian method \citep{kuhn1955hungarian}. Simpler sub-optimal alternatives also exist that operate with $\mathcal{O}\left(N^2\right)$ or $\mathcal{O}\left(N~\text{log}(N)\right)$ flops.

A question that naturally arises is how to generate powerful node embeddings that capture the network connectivity and also be effective in aligning different graphs. A natural and effective approach is to leverage the spectral decomposition of the graph, $\bm S = \bm V\bm\Lambda\bm V^T$,
where $\bm V$ is the orthonormal matrix of the eigenvectors, and $\bm \Lambda$ is the diagonal matrix of corresponding eigenvalues. Note that we assume undirected graphs and thus $\bm S$ is symmetric. Spectral decomposition has been proven to be an efficient approach to generating meaningful node embedding for graph matching \citep{6778, feizi2019spectral}. In particular, $\bm E =\bm V$ or $\bm E = \bm V\bm\Lambda$ are node embeddings that capture the network connectivity since they can perfectly reconstruct the graph. However, $\bm V$ is not unique. Thus computing the spectral decomposition of the same graph with node relabelling, $\tilde{\bm S} = \bm P{\bm S}\bm P^T$ is not guaranteed to produce a permuted version of $\bm V$, i.e., $\bm P\bm V$. Even in the case where $\bm S$ does not have repeated eigenvalues $\bm V$ is only unique up to column sign, which prevents effective matching.

To overcome the aforementioned uniqueness limitation, one can focus on the top $m$ eigenvectors that correspond to non-repeated eigenvalues in both $\bm S$ and $\hat{\bm S}$ and compute their absolute values. Then network alignment can be cast as:
\begin{equation}\label{eq:linearassignment2} \min_{\bm P\in\mathcal{P}}~\left\lVert~\left |\bm V_m\right |-\bm P\left |\hat{\bm V}_m\right |~\right\rVert_F^2,
\end{equation}
where $\bm V_m\in\mathbb{R}^{N\times m}$ corresponds to the subspace of non-repeated eigenvalues. The formulation in \eqref{eq:linearassignment2} is a similar to the problem solved in \citep{6778}.
\section{Graph Neural Networks (GNNs) Upper-Bounds Spectral Methods for Network Alignment}
A GNN is a cascade of layers and performs local, message-passing operations that are usually defined by the following recursive equation:
\begin{align}\label{eq:GNNrec00}
    x_v^{(l+1)} = g\left(x_v^{(l)},f\left(\left\{x_u^{(l)}:u\in\mathcal{N}\left(v\right)\right\}\right)\right),
\end{align}
where $\mathcal{N}\left(v\right)$ is the neighborhood of vertex $v$, i.e., $u\in\mathcal{N}\left(v\right)$ if and only if $(u,v)\in\mathcal{E}$. The function $f$ operates on multisets ($\{\cdot\}$ represents a multiset) and $f,~g$ are ideally injective. Common choices for $f$ are the summation or mean function, and for $g$ the linear function, or the multi-layer perceptron (MLP).

Overall, the output of the $L-$th layer of a GNN is a function $\phi\left(\bm X; \bm S, \mathcal{H} \right):\mathbb{R}^{N\times D}\to\mathbb{R}^{N\times D_{L}}$, where $\bm S$ is the graph operator, and $\mathcal{H}$ is the tensor of the trainable parameters in all $L$ layers and produces $D_{L}-$ dimensional embeddings for the nodes of the graph defined by $\bm S$.

GNNs admit some very valuable properties. First, they are permutation equivariant:
\begin{theorem}[\citep{xu2018how,maron2018invariant}]\label{thrm:equivariance}
Let $\phi\left(\bm X; \bm S, \mathcal{H} \right):\mathbb{R}^{N\times D}\to\mathbb{R}^{N\times D^{L}}$ be a GNN with parameters $\mathcal{H}$. For $\tilde{\bm X} = \bm P\bm X$ and $\tilde{\bm S} = \bm P\bm S\bm P^T$ that correspond to node relabelling according to the permutation matrix $\bm  P$, the output of the GNN takes the form:
\begin{equation}
    \tilde{\bm X}^{(L)} =\phi\left(\tilde{\bm X}; \tilde{\bm S}, \mathcal{H} \right)= \bm P\phi\left(\bm X; \bm S, \mathcal{H} \right)
\end{equation}
\end{theorem}
The above property is not satisfied by other spectral methods. GNNs are also stable \citep{gama2020stability, wang1, wang2}, transferable \citep{luana2020}, and have high expressive power \citep{xu2018how,abboud2020surprising,kanatsoulis2022graph}.
\subsection{GNNs and Network Alignment}\label{sec::theory}
To characterize the ability of a GNN to perform network alignment we first pointed out the GNNs perform nonlinear spectral operations. Details can be found in Appendix Section \ref{app:spectral}. We can further prove that:
\begin{theorem}\label{theorem:abs_eig}
Let $\mathcal{G},~\hat{\mathcal{G}}$ be graphs with adjacencies $\bm{S},~\hat{\bm{S}}$ that have non-repeated eigenvalues. Also
let ${\bm P^\diamond},~\check{\bm P}$ be solutions to the optimization problems in \eqref{eq:quadraticassignment} and \eqref{eq:linearassignment2} respectively. Then there exists a GNN $\phi\left(\bm X; \bm S, \mathcal{H} \right):\mathbb{R}^{N\times D}\to\mathbb{R}^{N\times D^{L}}$ such that:
\small{\begin{equation*}{\left\lVert~\bm S-\bm P^{\diamond}\hat{\bm S}\bm P^{\diamond^ T}\right\rVert_F^2\leq\left\lVert~\bm S-\bm P^{\ast}\hat{\bm S}\bm P^{\ast^ T}\right\rVert_F^2\leq\left\lVert~\bm S-\check{\bm P}\hat{\bm S}\check{\bm P}^T\right\rVert_F^2}
\end{equation*}}
\normalsize
with
\begin{equation*}
    \bm P^{\ast}=\argmin_{\bm P\in\mathcal{P}}~\left\lVert~\phi\left(\bm X; {\bm S}, \mathcal{H} \right)-\bm P\phi\left(\hat{\bm X}; \hat{\bm S}, \mathcal{H} \right)~\right\rVert_F^2.
\end{equation*}
\end{theorem}
The proof can be found in Appendix \ref{appendix:proof_of_theorem}. The assumption that the graph adjacencies have different eigenvalues is not restrictive. Real nonisomorphic graphs have different eigenvalues with very high probability \citep{haemers2004enumeration}. Theorem \ref{theorem:abs_eig} compares the network alignment power of a GNN with that of a spectral algorithm \cite{6778}, that uses the absolute values of graph adjacency eigenvectors to match two different graphs. According to Theorem \ref{theorem:abs_eig} there always exists a GNN that can perform at least as well as the spectral approach. The proof studies a GNN with white random input and measures the variance of the filter output. Then it shows that message-passing layers are able to compute the absolute values of the graph adjacency eigenvectors when the adjacency has non-repeated eigenvalues. As a result there always exists a single layer GNN that outputs the same node features as the ones used in \cite{6778}, which concludes our proof. The questions is: How do we train such a GNN that is (1) Expressive to the structural information thus its output can be used to match corresponding nodes. (2) Robust to different perturbations and even larger scale unseen graphs so that it can be deployed efficiently without re-training. (3) Agnostic to the ground truth node correspondence so that it can be trained unsupervisedly, which makes it generalizable to most real-world settings. To answer these questions, we introduce our proposed training framework in the following section.
\section{Proposed Method}\label{sec:: method}
 We now leverage the favorable properties of GNNs (permutation equivariance, expressivity, and transferability) to tackle real-world network alignment. Our approach learns low-dimensional node embeddings (Eq. \ref{eq:GNNrec00}) that enable graph matching via solving the linear assignment in \eqref{eq:linearassignment} rather than a quadratic assignment problem in \eqref{eq:quadraticassignment}. We design a robust GNN framework such that the node embeddings are expressive to accurately match similar nodes and also stable to graph perturbations.
\subsection{Learning Network Geometry with Transferable Graph Auto-encoders}
The goal of the proposed framework is to learn a function that maps graphs to node representations and effectively match nodes from different graphs. This function is modeled by a GNN encoder $\phi\left(\bm X; {\bm S}, \mathcal{H} \right)$, described by Fig. \ref{Model}. The learned encoder should work for a family of training graphs $\left\{\mathcal{G}_{0},\dots,\mathcal{G}_{i},\dots,\mathcal{G}_{I}\right\}$ with a set of adjacency matrices $\mathbb{S} = \left\{\bm{S}_{0},\dots,\bm{S}_{i},\dots,\bm{S}_{I}\right\}$, rather than a single graph. So the idea is not to train a GNN on a single graph \citep{GAE}, but train a transferable graph auto-encoder by solving the following optimization problem.
\begin{equation}\label{eq:family}\min_{\mathcal{H}}~\mathbb{E}\left[l\left(~\rho\left(\phi\left(\bm X; {\bm S_i}, \mathcal{H} \right)\phi\left(\bm X; {\bm S_i}, \mathcal{H} \right)^T\right), \bm S_i \right)\right],
\end{equation}
where $l\left(\cdot\right)$ is the binary cross entropy (BCE) and $\rho\left(\cdot\right)$ is the logistic function. $\bm S_i\in\mathbb{S}$ is a realization from a family of graphs and the expectation (empirical expectation is practice) is computed over this graph family. The generalized framework in \eqref{eq:family} learns a mapping from graphs to node representations, and can be applied to \textbf{\textit{out-of-distribution}} graphs that have not been observed during training. This twist in the architecture enables node embedding and graph matching for the unseen and larger scale networks without re-training, where fitting a GNN is computationally prohibitive in real-world applications.
\subsection{Robust and Generalizable Node representations with self-supervised learning (data augmentation)}
So far we proposed a GNN framework to produce expressive node representations to perform network alignment. In this subsection, we further upgrade our framework by ensuring the robustness and generalization ability of the proposed mapping. In particular, for each graph, $\bm{S}_{i} \in\mathbb{S}$, we augment the training set with perturbed versions that are described by the following set of graph adjacencies $\mathbb{M}_i=\left\{\bm{S}_{i}^{(0)},\dots,\bm{S}_{i}^{(j)},\dots,\bm{S}_{i}^{(J)}\right\}$, that are perturbed versions of $\bm S_i$. To do so we add or remove an edge with a certain probability yielding $\tilde{\bm S}_i\in\mathbb{M}$, such that $\tilde{\bm S}_i = \bm S_i + \bm M_i$, where $\bm M_i\in\{-1,0,1\}^{N\times N}$. Note that $\bm M$ changes for each $\tilde{\bm S}_i$, and $\bm M[m,n]$ can be equal to $1$ and $-1$ only if $\bm S[m,n]$ is equal to $0$ and $1$ respectively. To train the proposed transferable graph-autoencoder we consider the following optimization problem:
\begin{equation}\label{eq:trainingfinal}
\small
\min_{\mathcal{H}}~\mathbb{E}_{\mathbb{S}}\left[\mathbb{E}_{\mathbb{M}_i}\left[l\left(~\rho\left(\phi\left(\bm X; \tilde{\bm S}_i, \mathcal{H} \right)\phi\left(\bm X; \tilde{\bm S}_i, \mathcal{H} \right)^T\right), \bm S_i \right)\right]\right],
\end{equation}
where $\mathbb{E}_{\mathbb{S}}$ is the expectation with respect to the family of graphs $\mathbb{S}$ and $\mathbb{E}_{\mathbb{M}_i}$ is the expectation with respect to the perturbed graphs $\mathbb{M}_i$. In practice, $\mathbb{E}_{\mathbb{S}},~\mathbb{E}_{\mathbb{M}}$ correspond to empirical expectations. Note that training according to \eqref{eq:trainingfinal} also benefits the robustness of the model, which is crucial in deep learning tasks \citep{wang2022measure, he2021performance}. A schematic illustration of the training process can be found in Fig. \ref{TrainingPic}.
\begin{remark}(Large-scale network alignment by transferability of GNN)
\label{def:inj}

The proposed framework learns a mapping $\phi:\mathbb{G}\to\mathbb{R}^{N\times F}$ that produces expressive and robust node representations for a family of graphs $\mathcal{G}\in\mathbb{G}$. This mapping is designed in such a way that the problem in \eqref{eq:linearassignment} approximates the problem in \eqref{eq:quadraticassignment} and allows solving network alignment in polynomial time. One of the main benefits of the proposed framework is that it enables large-scale network alignment. Task specific augmentation during training is the key to prompt transferability of deep neural networks \citep{augment}. And the transferability analysis of GNN encoders \citep{luana2020} suggests that we can train with small graphs and efficiently execute with much larger graphs when the substructures (motifs) that appear in the tested graphs, were also partially observed during training. Since the proposed transferable graph auto-encoder is trained with multiple graphs, a variety of motifs are observed during training, which cannot be observed with a classical graph autoencoder, and the proposed GNN encoder can be transferred to larger-scale out-of-distribution graphs. 
\end{remark}
\vspace{-0.2cm}
\subsection{Alignment and Complexity analysis}\label{sec::alignment_complexity}
\begin{figure}
        \centering
         \includegraphics[height=4cm, width=\linewidth]{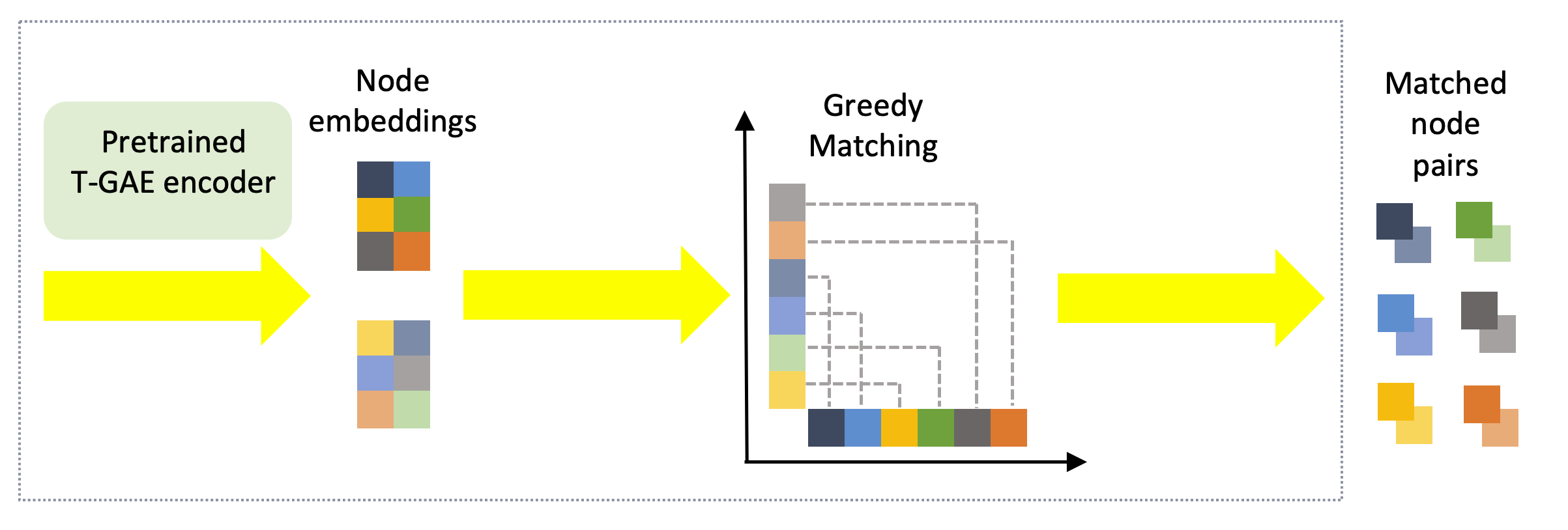}
    \caption{The pre-trained encoder operates on out-of-distribution samples. The generated node embeddings are then matched greedily.}
\label{inferencePic}
\vspace{-0.4cm}
\end{figure}
After learning the powerful node embeddings, network alignment is performed by solving the linear assignment problem in \eqref{eq:linearassignment}. An illustration of the assignment is presented in Fig. \ref{inferencePic}. The node features produced by
T-GAE are used to calculate a pairwise distance
matrix, followed by the greedy Hungarian algo
rithm to predict node correspondences. To analyze the complexity of our approach we study the 3 main parts of T-GAE: a) The design of the input structural features, b) The message-passing GNN that produces node embeddings, and c) the linear assignment algorithm.

The computation of our neighborhood-based structural features is expected to take $\mathcal{O}\left(\left|\mathcal{V}\right|\right)$ in real graphs, as proved in \cite{henderson2011s}. The computational and memory complexity of the message-passing GNN is $\mathcal{O}\left(\left|\mathcal{V}\right|c^2+\left|\mathcal{E}\right|c\right)$, and $\mathcal{O}\left(\left|\mathcal{V}\right|c\right)$, where $c$ is the width of the GNN. The computational complexity to align the nodes of the graph is $\mathcal{O}\left(\left|\mathcal{V}\right|^2\right)$ since we are using the suboptimal greedy Hungarian. If we want to optimally solve the linear assignment problem we need to use the Hungarian algorithm that has $\mathcal{O}\left(\left|\mathcal{V}\right|^3\right)$ complexity. If we want to process large graphs we can use efficient nearest neighbors algorithms with complexity $\mathcal{O}\left(\left|\mathcal{V}\right|\text{log}\left(\left|\mathcal{V}\right|\right)\right)$ to perform soft linear assignment. However, this efficient algorithm only works to match graphs with its permuted samples. We include detailed discussion in Appendix Section \ref{sec::efficiency}. Overall the complexity of T-GAE is 
$\mathcal{O}\left(\left|\mathcal{V}\right|^2\right)$, or $\mathcal{O}\left(\left|\mathcal{V}\right|c^2+\left|\mathcal{E}\right|c+\left|\mathcal{V}\right|\text{log}\left(\left|\mathcal{V}\right|\right)\right)$ for un-perturbed samples.

\section{Experiments}\label{sec::main_exp}
\subsection{Experiments Setup}
The experiments included in this section are designed to answer the following research questions:(1) Can \texttt{T-GAE} generate competing graph matching accuracy on real-world networks from various domains with different sizes, while being efficient by utilizing transferability of GNN for large-scale our-of-distribution graphs? We answer this question in Section \ref{sec::main_exp} and Appendix Section \ref{section:degreeModel} by comparing the performance on matching small to middle sized graphs with unseen perturbed samples, and large scale out-of-distribution networks under different perturbation distributions. (2) Is \texttt{T-GAE} robust to different graph matching tasks and real-world noise distributions? In Section \ref{sec::subgraph}, we conduct sub-graph matching experiments to align two different real-world networks that are partially aligned. (3) How does the proposed network architecture (Figure \ref{Model}) and the training objective (Figure \ref{TrainingPic}) contribute to network alignment? To demonstrate the contribution of each proposed component, we conduct ablation studies in Section \ref{sec::ablation}, to compare T-GAE with untrained T-GAE, T-GAE trained with GAE objective, and GAE. (4) How much efficiency does T-GAE offer compared to the existing optimization and GNN approaches, and what are the possible trade-offs between efficiency and matching accuracy of T-GAE? We compare the efficiency of different matching algorithms and empirically prove that (a) The matching accuracy of T-GAE can be further improved by leveraging the exact Hungarian algorithm. (b) The efficiency of T-GAE to match graphs with their permuted samples can be enhanced if we adopt the more efficient matching algorithm introduced in Section \ref{sec::alignment_complexity}. This set of experiments are included in Appendix Section \ref{sec::efficiency}.

For each of the above mentioned experiments, we compare T-GAE with three categories of graph matching approaches: (a)GNN based methods: \texttt{WAlign} \citep{jundong}, \texttt{GAE} and \texttt{VGAE} \citep{GAE}; (b)Graph/Node embedding techniques: \texttt{NetSimile} \citep{netsimile}, \texttt{Spectral} \citep{6778}, \texttt{DeepWalk} \citep{deepwalk}, \citep{node2vec}, \texttt{GraphWave} \citep{graphwave} and \texttt{LINE} \citep{line}. {(c)}Optimization based graph matching algorithms: \texttt{S-GWL} \citep{sgwl}, \texttt{ConeAlign} \citep{chen2020cone} and \texttt{FINAL} \citep{onlineoffline}. Note that \texttt{LINE, VGAE, DeepWalk}, and \texttt{Node2Ve}c are omitted from some experiments since they show very poor performance. The reason behind that is that they are not permutation equivariant. GraphWave is also excluded from the sub-graph matching experiment, it could not identify correlated nodes in two different graphs. In the case of graphs without attributes \texttt{FINAL} is equivalent to the popular \texttt{Isorank} \citep{bio2} algorithm, and \texttt{FINAL} is omitted in sub-graph matching experiments due to weak performance. 

We include detailed descriptions of our included datasets, implementation details of T-GAE and all the competing baselines in Appendix Section \ref{section:impSup}. 

\subsection{Graph Matching Experiments}\label{section:pert}
In this subsection we compare the performance of T-GAE with all competing baselines to match the graphs with permuted and perturbed versions of them. In particular, let $\mathcal{G}$ be a graph with adjacency matrix $\bm S$. We then produce 10 permuted-perturbed versions according to $\hat{\bm S} = \bm P\left(\bm S + \bm M\right)\bm P^T$, where $\bm M\in\{-1,0,1\}^{N\times N}$ and $\bm P$ is a permutation matrix. For each perturbation level $p \in \{0, 1\%, 5\%\}$, the total number of perturbations is defined as $p|\mathcal{E}|$, where $|\mathcal{E}|$ is the number of edges of the original graph.

Specifically, we train \texttt{T-GAE} according to \eqref{eq:trainingfinal}, where $\mathbb{S}$ consist of the small-size networks, i.e., Celegans, Arena, Douban, and Cora. Then we resort to transfer learning and use the \texttt{T-GAE} encoder to produce node embedding for perturbed versions of (a) Celegans, Arena, Douban, and Cora, and (b) larger graphs, i.e., Dblp, and Coauthor CS. Note that none of these perturbed versions were considered during training. This is in contrast with all GNN baselines that are retrained on every pair of networks in the testing dataset. We report the average and standard deviation of the matching accuracy for 10 randomly generated perturbation samples under uniform edge editing in Table \ref{tab:transferuniform}, where each edge and non-edge shares the same probability of being removed or augmented. We report the results for removing edges according to degrees and the relevant discussed in Appendix \ref{section:degreeModel}.
\begin{table*}[t]
\centering
  \caption{Graph matching accuracy on 10 randomly perturbed samples under different levels of edge editing on Uniform model. The proposed T-GAE is trained on the clean Celegans, Arena, Douban, and Cora networks, and tested on noisy versions of them and the larger Dblp, and Coauthor CS. Accuracy above 80\% is highlighted in green, 60\% to 80\% accuracy is in yellow, and performance below 60\% is in red.}\label{tab:transferuniform}
  \small
\setlength\tabcolsep{2pt}
\scalebox{0.7}{\begin{tabular}{c|P{2.7cm}|P{1.7cm}|P{1.7cm}|P{1.7cm}|P{1.7cm}|P{1.7cm}|P{1.7cm}|P{1.7cm}|P{1.7cm}|P{1.7cm}}
    \toprule
      &\multirow{2}{*}{Dataset $\backslash$ Algorithm}&
      \multicolumn{3}{c|}{\textit{\textbf{Feature Engineering based}}} &
      \multicolumn{3}{c|}{\textit{\textbf{Optimization based}}} &
      \multicolumn{3}{c}{\textit{\textbf{GNN based}}} \\
   &&{Spectral}&{Netsimile}&{GraphWave}&{FINAL}&{S-GWL}&{ConeAlign}&{WAlign}&{GAE}&\textbf{T-GAE}\\
    \midrule
     {\multirow{6}{*}{\rotatebox[origin=c]{90}{$0\%$ perturbation}}}& \cellcolor{white}Celegans & \cellcolor{green!50} ${87.8\pm1.5}$ & \cellcolor{yellow!50} ${72.7\pm 0.9}$ & \cellcolor{yellow!50} ${65.3\pm1.7}$ & \cellcolor{green!50} ${92.2\pm 1.2}$& \cellcolor{green!50}$\mathbf{93.0\pm1.5}$ & \cellcolor{yellow!50} ${66.6\pm1.2}$ &\cellcolor{green!50}${88.4\pm1.6}$ &\cellcolor{green!50}${86.3\pm1.3}$ &\cellcolor{green!50}${91.0\pm1.1}$\\
     
     &\cellcolor{white}Arenas & \cellcolor{green!50}${97.7\pm0.4}$ &\cellcolor{green!50}${94.7\pm0.3}$  &\cellcolor{green!50}${81.7\pm0.7}$ &\cellcolor{green!50}${97.5\pm0.3}$&\cellcolor{green!50}${97.5\pm0.3}$  &\cellcolor{green!50}${87.8\pm0.6}$&\cellcolor{green!50}${97.4\pm0.5}$ &\cellcolor{green!50}${97.6\pm0.4}$ &\cellcolor{green!50}$\mathbf{97.8\pm0.4}$\\

    &\cellcolor{white}Cora &\cellcolor{green!50}${85.0\pm0.4}$ &\cellcolor{yellow!50}${73.7\pm0.4}$ &\cellcolor{red!50}${8.3\pm0.4}$&\cellcolor{green!50}${87.5\pm0.7}$ &\cellcolor{green!50}${87.3\pm0.7}$ &\cellcolor{red!50}${38.5\pm0.7}$ &\cellcolor{green!50}${87.2\pm0.4}$ &\cellcolor{green!50}${87.1\pm0.8}$ &\cellcolor{green!50}$\mathbf{87.5\pm0.4}$\\
          
     &\cellcolor{white}Douban &\cellcolor{green!50}${89.9\pm0.4}$ &\cellcolor{yellow!50}${46.4\pm0.4}$  &\cellcolor{red!50}${17.5\pm0.2}$ &\cellcolor{green!50}${89.9\pm0.3}$&\cellcolor{green!50}$\mathbf{90.1\pm0.3}$ &\cellcolor{yellow!50}${68.1\pm0.4}$ &\cellcolor{green!50}${90.0\pm0.4}$ &\cellcolor{green!50}${89.5\pm0.4}$ &\cellcolor{green!50}$\mathbf{90.1\pm0.3}$\\
     
     &\cellcolor{white}Dblp &\cellcolor{green!50}${84.5\pm0.1}$ &\cellcolor{yellow!50}${63.7\pm0.2}$  &\cellcolor{red!50}doesn't scale &\cellcolor{green!50}$\mathbf{85.6\pm0.2}$&\cellcolor{red!50}> 48 hours &\cellcolor{red!50}${44.3\pm0.6}$ &\cellcolor{green!50}${85.6\pm0.2}$ &\cellcolor{green!50}${85.2\pm0.3}$  &\cellcolor{green!50}$\mathbf{85.6\pm0.2}$\\
     
     &\cellcolor{white}Coauthor CS &\cellcolor{green!50}${97.5\pm0.1}$ &\cellcolor{green!50}${90.9\pm0.1}$ &\cellcolor{red!50}doesn't scale &\cellcolor{green!50}$\mathbf{97.6\pm0.1}$ &\cellcolor{red!50}> 48 hours &\cellcolor{yellow!50}${75.8\pm0.5}$ &\cellcolor{green!50}${97.5\pm0.2}$ &\cellcolor{green!50}${97.6\pm0.3}$ &\cellcolor{green!50}$\mathbf{97.6\pm0.1}$\\
     
     \midrule
     {\multirow{6}{*}{\rotatebox[origin=c]{90}{$1\%$ perturbation}}}& \cellcolor{white}Celegans & \cellcolor{yellow!50} ${68.5\pm16.1}$ & \cellcolor{yellow!50} ${66.3\pm 3.8}$  & \cellcolor{red!50} ${22.5\pm22.4}$ & \cellcolor{red!50} ${33.2\pm 7.8}$& \cellcolor{green!50}$\mathbf{87.1\pm6.1}$ & \cellcolor{yellow!50} ${60.9\pm2.5}$ &\cellcolor{green!50}${80.7\pm3.0}$ &\cellcolor{red!50}${33.2\pm8.4}$ &\cellcolor{green!50}${86.5\pm1.1}$\\
     
     &\cellcolor{white}Arenas & \cellcolor{green!50}${85.0\pm10.0}$ &\cellcolor{green!50}${87.8\pm1.0}$ &\cellcolor{red!50}${40.5\pm23.8}$ &\cellcolor{red!50}${32.5\pm5.9}$&\cellcolor{green!50}${94.2\pm0.7}$ &\cellcolor{green!50}${84.6\pm1.0}$ &\cellcolor{green!50}${90.0\pm3.1}$ &\cellcolor{red!50}${30.1\pm17.6}$ &\cellcolor{green!50}$\mathbf{96.0\pm1.0}$\\

    &\cellcolor{white}Cora &\cellcolor{red!50}${59.1\pm9.3}$ &\cellcolor{yellow!50}${66.4\pm1.6}$ &\cellcolor{red!50}${3.7\pm2.9}$ &\cellcolor{red!50}${30.0\pm3.3}$&\cellcolor{red!50}${46.4\pm6.9}$ &\cellcolor{red!50}${33.5\pm1.6}$ &\cellcolor{green!50}${80.1\pm1.2}$ &\cellcolor{red!50}${57.9\pm5.3}$ &\cellcolor{green!50}$\mathbf{85.1\pm0.5}$\\
          
     &\cellcolor{white}Douban &\cellcolor{red!50}${25.8\pm27.2}$ &\cellcolor{red!50}${40.0\pm1.2}$ &\cellcolor{red!50}${9.9\pm5.9}$ &\cellcolor{red!50}${27.8\pm5.7}$ &\cellcolor{yellow!50}${72.1\pm0.7}$ &\cellcolor{yellow!50}${64.7\pm0.4}$ &\cellcolor{yellow!50}${77.2\pm4.8}$ &\cellcolor{red!50}${38.3\pm16.4}$ &\cellcolor{green!50}$\mathbf{87.3\pm0.4}$\\
     
     &\cellcolor{white}Dblp &\cellcolor{red!50}${55.6\pm19.0}$ &\cellcolor{red!50}${55.1\pm1.7}$ &\cellcolor{red!50}doesn't scale &\cellcolor{red!50}${15.2\pm3.3}$ &\cellcolor{red!50}> 48 hours &\cellcolor{red!50}${37.8\pm1.1}$ &\cellcolor{yellow!50}${73.1\pm1.6}$ &\cellcolor{red!50}${19.4\pm0.6}$  &\cellcolor{green!50}$\mathbf{83.3\pm0.4}$\\
     
     &\cellcolor{white}Coauthor CS &\cellcolor{red!50}${58.2\pm22.1}$ &\cellcolor{yellow!50}${75.2\pm2.2}$   &\cellcolor{red!50}doesn't scale &\cellcolor{red!50}${13.3\pm5.0}$ &\cellcolor{red!50}> 48 hours &\cellcolor{yellow!50}${68.5\pm2.8}$ &\cellcolor{yellow!50}${75.2\pm5.4}$ &\cellcolor{red!50}${49.5\pm7.8}$ &\cellcolor{green!50}$\mathbf{93.2\pm0.8}$\\
     \midrule
     {\multirow{6}{*}{\rotatebox[origin=c]{90}{$5\%$ perturbation}}}& \cellcolor{white}Celegans & \cellcolor{red!50} ${24.9\pm15.9}$ & \cellcolor{red!50} ${41.1\pm 13.0}$ & \cellcolor{red!50} ${7.6\pm9.2}$ & \cellcolor{red!50} ${10.4\pm 2.7}$ & \cellcolor{yellow!50}${68.3\pm12.7}$ & \cellcolor{red!50} ${50.5\pm3.4}$ &\cellcolor{red!50}${42.4\pm21.1}$ &\cellcolor{red!50}${6.5\pm2.4}$ &\cellcolor{yellow!50}$\mathbf{69.2\pm2.1}$\\
     
     &\cellcolor{white}Arenas & \cellcolor{red!50}${52.1\pm16.5}$ &\cellcolor{red!50}${52.3\pm5.3}$   &\cellcolor{red!50}${6.9\pm7.2}$ &\cellcolor{red!50}${7.2\pm2.6}$ &\cellcolor{green!50}$\mathbf{88.3\pm3.2}$ &\cellcolor{yellow!50}${75.0\pm2.7}$ &\cellcolor{red!50}${30.4\pm17.5}$ &\cellcolor{red!50}${1.4\pm1.4}$ &\cellcolor{green!50}${81.2\pm1.4}$\\

     &\cellcolor{white}Cora &\cellcolor{red!50}${29.5\pm0.8}$ &\cellcolor{red!50}${41.2\pm3.3}$ &\cellcolor{red!50}${0.8\pm0.3}$ &\cellcolor{red!50}${6.7\pm2.8}$ &\cellcolor{red!50}${39.9\pm5.5}$ &\cellcolor{red!50}${23.0\pm2.0}$&\cellcolor{red!50}${33.4\pm7.3}$ &\cellcolor{red!50}${9.6\pm2.7}$ &\cellcolor{yellow!50}$\mathbf{67.7\pm1.3}$\\
     
     &\cellcolor{white}Douban &\cellcolor{red!50}${23.8\pm20.6}$ &\cellcolor{red!50}${20.7\pm4.6}$ &\cellcolor{red!50}${1.9\pm2.8}$ &\cellcolor{red!50}${7.8\pm3.0}$ &\cellcolor{yellow!50}${68.6\pm0.8}$ &\cellcolor{yellow!50}${54.1\pm1.2}$ &\cellcolor{red!50}${36.6\pm13.4}$ &\cellcolor{red!50}${0.6\pm0.3}$ &\cellcolor{yellow!50}$\mathbf{70.2\pm2.5}$\\
     
     &\cellcolor{white}Dblp &\cellcolor{red!50}${28.0\pm7.8}$ &\cellcolor{red!50}${19.5\pm4.8}$ &\cellcolor{red!50}doesn't scale &\cellcolor{red!50}${2.7\pm0.9}$  &\cellcolor{red!50}> 48 hours &\cellcolor{red!50}${24.4\pm2.9}$ &\cellcolor{red!50}${15.9\pm8.3}$ &\cellcolor{red!50}${1.4\pm0.2}$  &\cellcolor{yellow!50}$\mathbf{60.8\pm1.9}$\\
     
     &\cellcolor{white}Coauthor CS &\cellcolor{red!50}${9.7\pm5.0}$ &\cellcolor{red!50}${26.3\pm6.0}$ &\cellcolor{red!50}doesn't scale &\cellcolor{red!50}${2.0\pm0.4}$  &\cellcolor{red!50}> 48 hours &\cellcolor{red!50}${51.4\pm5.1}$ &\cellcolor{red!50}${11.3\pm7.5}$ &\cellcolor{red!50}${0.6\pm0.1}$ &\cellcolor{yellow!50}$\mathbf{66.0\pm1.4}$\\
    \bottomrule
\end{tabular}}
\end{table*}
\begin{table*}[t]
\centering
  \caption{Subgraph matching performance comparison. The proposed T-GAE is trained on the two real-world graphs, and test to match the aligned portion of them.}
  \small
\setlength\tabcolsep{2pt}
\scalebox{0.85}{\begin{tabular}{|c|P{1.5cm}|P{1.5cm}|P{1.5cm}|P{1.5cm}|P{1.5cm}|P{1.5cm}|P{1.5cm}|P{1.5cm}|}
    \toprule
      \multirow{2}{*}{Algorithm $\backslash$ Hit Rate}&
      \multicolumn{4}{c|}{\textit{\textbf{ACM-DBLP}}} &
      \multicolumn{4}{c|}{\textit{\textbf{Douban Online-Offline}}} \\
   &{Hit@1}&{Hit@5}&{Hit@10}&{Hit@50}&{Hit@1}&{Hit@5}&{Hit@10}&{Hit@50}\\
    \midrule
    {Netsimile}&{2.59\%}&{8.32\%}&{12.09\%}&{26.42\%}&{1.07\%}&{2.77\%}&{4.74\%}&{15.03\%}\\
    {Spectral} &{1.40\%}&{4.62\%}&{7.21\%}&{16.34\%}&{0.54\%}&{1.34\%}&{2.95\%}&{13.95\%}\\
    {GAE}&{8.1\%}&{22.5\%}&{30.1\%}&{45.1\%}&{3.3\%}&{9.2\%}&{14.1\%}&{32.1\%}\\
    {$\text{WAlign}$}&{62.02\%}&{81.96\%}&{87.31\%}&{93.89\%}&{${36.40\%}$}&{${53.94\%}$}&{${67.08\%}$}&{85.33\%}\\
    {$\text{T-GAE}$}&{$\mathbf{73.89\%}$}&{$\mathbf{91.73\%}$}&{$\mathbf{95.33\%}$}&{$\mathbf{98.22\%}$}&{$\mathbf{36.94\%}$}&{$\mathbf{60.64\%}$}&{$\mathbf{69.77\%}$}&{$\mathbf{89.62\%}$}\\
    \bottomrule
\end{tabular}}
\label{subgraph}
\end{table*}

\textbf{T-GAE framework results in a robust and transferable GNN to perform network alignment at a large scale.} Our first observation is that for zero perturbation most algorithms are able to achieve a high level of matching accuracy. This is expected, since for zero perturbation the network alignment is equivalent to graph isomorphism. On the smaller networks(Celegans, Arenas, Douban, Cora), \texttt{T-GAE} performs at least as well as the current state-of-the-art optimization approaches (\texttt{S-GWL} and \texttt{ConeAlign}). Specifically, it achieves up to 38.7\% and 44.7\% accuracy increase compared to \texttt{S-GWL} and \texttt{ConeAlign}, respectively. Regarding the ability of \texttt{T-GAE} to perform large-scale network alignment the results are definitive. \texttt{T-GAE} enables low-complexity training with small graphs, and execution at larger settings by leveraging transfer learning, and it consistently outperforms all competing baselines on the two networks with more than 10k nodes. In particular, it is able to achieve very high levels of matching accuracy for both Dblp and Coauthor CS, for $p=0\%,~1\%$. It is also the only method that consistently achieves at least $60\%$ accuracy at $5\%$ perturbation. To the best of our knowledge, our experiments on DBLP \citep{PanWZZW16} and Coauthor CS \citep{shchur2018pitfalls} are the first attempts to perform exact alignment on networks at the order of $20$k nodes and $80$k edges.

\textbf{The benefits of processing structural node features with T-GAE is clear.} There is a clear benefit of processing the structural embeddings with \texttt{TGAE} since it offers up to $43.7\%$ performance increase compared to \texttt{NetSimile}. When some perturbation is added, the conclusions are straightforward. Especially when perturbations are added, our proposed \texttt{T-GAE} markedly outperforms all the competing GNN alternatives(\texttt{WAlign} and \texttt{GAE}) and shows the desired robustness to efficiently perform network alignment. We observe that neither GAE nor WAlign is robust to noise. This highlights the benefit of T-GAE in handling the distribution shift brought by the structural dissimilarity between different graphs. 
\subsection{Sub-graph Matching Experiments}\label{sec::subgraph}
We further test the performance of T-GAE in matching subgraphs of different networks that have aligned nodes (nodes that represent the same entities in different networks). Specifically, in ACM-DBLP, the task is to match the papers that appear in both citation networks; in Douban Online-Offline, we aim to identify the users that take part into both online and offline activities. Note that this is the most realistic graph matching experiment we can perform since we match 2 different real graphs with partially aligned nodes.

\textbf{T-GAE uniformly achieves the best performance in the most realistic scenario of network alignment.} Most optimization based approaches do not generalize to this real-world scenario because their optimization objective usually prevents from matching graphs with different numbers of nodes. There is a significant improvement in matching accuracy with GNN-based methods (TGAE and WAlign) compared to traditional graph or node embedding techniques. In particular, \texttt{T-GAE} consistently achieves the best performance among all competing methods. This suggests that the encoder model illustrated in Fig. \ref{Model} and the training framework (\ref{eq:trainingfinal}) illustrated in Fig. \ref{TrainingPic}, provide an efficient approach to generate powerful node embedding, that is robust to real-world noise distributions for the task of network alignment, compared to the existing GNN frameworks.
\subsection{Ablation study} \label{sec::ablation}
\begin{figure}
        \centering
         \includegraphics[height=5cm, width=\linewidth]{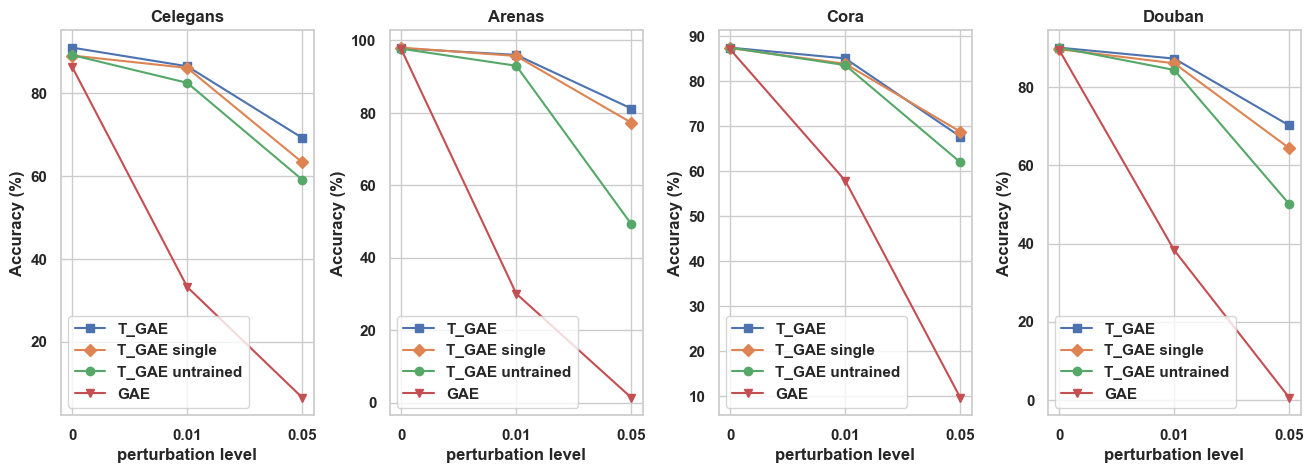}
    \caption{Graph matching performance comparison of T-GAE,T-GAE trained on a single graph(T-GAE single), the untrained T-GAE (T-GAE untrained), and GAE. The proposed training objective and encoder structure helps to prompt the expressiveness of GNN thus achieve higher accuracy as we introduce more perturbations.}
\label{ablationPic}
\vspace{-0.5cm}
\end{figure}
\textbf{The proposed architecture and training objective prompts the robustness of GNN when matching graphs with their highly perturbed versions.} From Figure \ref{ablationPic}, we observe that T-GAE outperforms the untrained T-GAE by a great margin when matching highly permuted samples. This implies that the proposed training objective effectively improves the robustness of GNN, which is the key property to deploy GNN to match perturbed graphs. Further, the performance gap between GAE and T-GAE single underscores the efficacy of incorporating attention mechanism on each layer for both input and output node features, as illustrated in Figure \ref{Model}.

\section{Limitation}
Although our approach achieves state-of-the-art performance in aligning real-graphs, on both graph matching and sub-graph matching tasks, approaching network alignment with a learning method, remains a heuristic and does not offer optimality guarantees. Furthermore, in order to process large graphs we cast network alignment as a self-supervised task. As a result in small-scale settings where the task can be tackled with computationally intensive efficient methods, our algorithm is not expected to perform the best. Finally, the complexity of T-GAE $\mathcal{O}(|\mathcal{V}|^2)$ is limiting, this bottleneck comes from the greedy linear assignment algorithm to match the node embedding, and therefore the alternative method with 
complexity $\mathcal{O}(|\mathcal{V}|c^2+|\mathcal{E}|c+|\mathcal{V}|\text{log}(|\mathcal{V}|))$ should be deployed when we match very large scale graphs with their permuted versions.

\section{Conclusion}
We proposed T-GAE, a graph autoencoder framework that utilizes transferability and robustness of GNN to perform network alignment. T-GAE is an unsupervised approach that tackles the high computational cost of existing optimization based algorithms, and can be trained on multiple small to middle sized graphs to produce robust and permutation equivariant embeddings for larger scale unseen networks. We proved that the produced embeddings of GNNs are related to the spectral decomposition of the graph and are at least as good in graph matching as certain spectral methods. Our experiments with real-world benchmarks on both graph matching and sub-graph matching demonstrated the great potential of utilizing the good properties of GNNs to solve network optimization problems in a more efficient and scalable way.
\bibliographystyle{unsrtnat}
\bibliography{reference}
\clearpage
\appendix
\section{Appendix}
\section{Notation}
Our notation is summarized in Table \ref{tab:TableOfNotationForMyResearch}.
\begin{table}[ht]
\setcounter{table}{0}
\caption{Key notations used in this paper.}
	\centering 
	{\small\begin{tabular}{r c p{6cm} }
		\toprule
		$\mathcal{G}$ & $\triangleq$ & Graph\\
		$\mathcal{V}$ & $\triangleq$ & Set of nodes\\
		$\mathcal{E}$ & $\triangleq$ & Set of edges\\
            $N$           & $\triangleq$ & Number of nodes\\
            $\bm{D}$           & $\triangleq$ & Degree matrix\\
		$\bm{S}$ & $\triangleq$ & $\{0,1\}^{N \times N}$ adjacency matrix\\	
		$\bm{X}$ & $\triangleq$ & $N \times D$ feature matrix\\
            $\bm{H}$ & $\triangleq$ & aggregation results of GNN convolution\\
            $\bm{W}$  & $\triangleq$   & weight matrix of the Graph Neural Network\\
            $\bm{N}_{v}$ & $\triangleq$ &neighbors of node v\\
            $\bm{I}$& $\triangleq$ &  Identity matrix\\
		$\bm{0}$& $\triangleq$ &  vector or matrix of zeros\\
		$\bm{A}^T$ & $\triangleq$ & transpose of matrix $\bm{A}$\\	
		$\bm{A}_{rc}$ & $\triangleq$ & entry at r-th row and j-th column of matrix $\bm{A}$\\
  $\lVert\cdot\rVert_F$ & $\triangleq$ & Frobenius norm\\

		\bottomrule
	\end{tabular}}
	\label{tab:TableOfNotationForMyResearch}
\end{table}
\section{Network Alignment in broader domains}
In this paper, we study the general form of network alignment to match nodes between two graphs. This task is challenging because of the little given information. However, it's not trivial to correctly and efficiently utilize the information contained in text-rich networks, such as KGs \citep{give}. In the domain of knowledge graph (KG) mining, entity alignment (EA) is the task to identify the same entities existing in knowledge graphs. \citep{EA1} proposes to replace the labor-intensive pre-processing with entity names mining. A structural-based refinement procedure is then applied to refine the entity name matching results. \citep{EA2, EA4} solves large scale EA  by aligning the KG in mini-batches. \citep{EA3} further proposes a self-supervised EA framework by automatically generating positive and negative matched node pairs. \citep{EA5} generalizes the unsupervised matching algorithm to temporal KGs. Specifically, it encodes the temporal and relational information respectively before an innovative jointly decoding process. Recently, the ability to deploy EA algorithms in real-world scenarios is enhanced by \citep {EA6}, which aligns multi-modal knowledge graphs.

Network alignment, as an important problem, has been studied not only by the community of data mining, it has also been mathematically and statistically investigated. A number of approaches have been proposed to solve this problem for Erd\H{o}s R\'enyi random graphs $G(n,\frac{d}{n})$. It has been proved that a perfectly true vertex correspondence can be recovered in polynomial time with high probability \citep{stat1}. Furthermore, a sharp threshold has been proved for both  Erd\H{o}s R\'enyi model and Gaussian model \citep{stat2}. Most recently, a novel approach to calculate similarity scores based on counting weighted trees rooted at each vertex has been proposed \cite{stat3}. Such approach has been proved to be effective in solving the aforementioned network alignment problem on random graphs with high probability. Readers are encouraged to refer to the authors of these publications \citep{stat1, stat2,stat3} for further reading. 

\section{Proof of Theorem \ref{theorem:abs_eig}} \label{appendix:proof_of_theorem}
\subsection{Spectral characterization of GNNs}\label{app:spectral}
What remains to be answered is the ability of a GNN to approximate a function that performs graph alignment. To understand the function approximation properties of GNNs we study them in the spectral domain. To this end, we consider the recursive formula in \eqref{eq:GNNrec00} where $f$ is the summation function and $g$ is multivariate linear for $K-2$ layers, and the MLP in the $(K-1)$-th layer. The overall operation can be written in a matrix form as:
\begin{equation}\label{eq:GNNreclinear}
    \bm X^{(l+1)} = \sigma\left(\sum_{k=0}^{K-1}\bm S^k \bm X^{(l)}\bm H_k^{(l)}\right),
\end{equation}
where $\bm H_k^{(l)}\in\mathbb{R}^{D^{l+1}\times D^{l}}$ is a linear mapping. Computing the spectral decomposition of $\bm S$ yields:
\begin{equation}\label{eq:GNNreclinear2}
    \bm X^{(l+1)} = \sigma\left(\sum_{k=0}^{K-1}\bm V\bm \Lambda^k\bm V^T \bm X^{(l)}\bm H_k^{(l)}\right)= \sigma\left(\sum_{k=0}^{K-1}\sum_{n=1}^{N}\lambda_n^k\bm v_n \bm v_n^T \bm X^{(l)}\bm H_k^{(l)}\right).
\end{equation}
Then each each column of $\bm X^{(l+1)}$ can be written as
\begin{align}\label{eq:GNNreclinearelement}
    \bm X^{(l+1)}[:,i] &= \sigma\left(\sum_{k=0}^{K-1}\sum_{n=1}^{N}\lambda_n^k\bm v_n \bm v_n^T \bm X^{(l)}\bm H_k^{(l)}[:,i]\right)=\sigma\left(\sum_{n=1}^{N}a_n^{(i)}\bm v_n\right),
\end{align}
where $\lambda_n,\bm v_n$ are the $n-$th eigenvalue and eigenvector and $a_n^{(i)}=\bm v_n^T \bm X^{(l)}\sum_{k=0}^{K-1}\lambda_n^k\bm H_k^{(l)}[:,i]$ is a scalar related to the Graph Fourier Transform (GFT) of $\bm X^{(l)}$ \citep{sardellitti2017graph}. It is clear from equation \eqref{eq:GNNreclinearelement} that the output of each layer is a linear combination of the adjacency eigenvectors, followed by a pointwise non-linearity. Thus, a GNN can produce unique and more powerful graph embeddings than spectral methods by processing the eigenvectors and eigenvalues of the adjacency matrix.
To prove Theorem \ref{theorem:abs_eig}. We consider one layer GNN with a vector input $\bm x\in\mathbb{R}^{N}$. This GNN can be represented by the following equation:
\begin{equation}\label{eq:GFouter}
    \bm Y = \sigma\left(\sum_{k=0}^{K-1}\bm S^k \bm x \bm h_k^T\right),
\end{equation}
where $\bm h_k\in\mathbb{R}^{m}$ and $\bm x \bm h_k^T$ is an outer-product operation. The equation in \eqref{eq:GFouter} describes a set of $m$ graph filters of the form:
\begin{equation}\label{eq:GFelement}
    \bm y_i = \sigma\left(\sum_{k=0}^{K-1}h_k^{i}\bm S^k \bm x\right),~~\text{for}~i=1,\dots, m
\end{equation}
\subsection{White random input and variance computation}
Let $\bm x$ be a white random vector with $\mathbb{E}\left[\bm x\right] = 0$ and $\mathbb{E}\left[\bm x\bm x^T\right] = \bm I$, where $\bm I$ is the diagonal matrix. Also let $\sigma\left(\cdot\right) = \left(\cdot\right)^2$ be the elementwise square function. Then \eqref{eq:GFelement} can be written as:
\begin{equation}
    \bm y_i = \left(\sum_{k=0}^{K-1}h_k^{i}\bm S^k \bm x\right)^2 = \text{diag}\left(\sum_{k=0}^{K-1}h_k^{i}\bm S^k \bm x\bm x^T\sum_{j=0}^{K-1}h_j^{i}\bm S^{j^T} \right)
\end{equation}
Since $\bm x$ is a random vector $\bm y_i$ is also a random vector. The expected value of $\bm y_i$ yields:
\begin{align}\label{eq:expectationproof}
    \mathbb{E}\left[\bm y_i\right] &= \mathbb{E}\left[ \text{diag}\left(\sum_{k=0}^{K-1}h_k^{i}\bm S^k \bm x\bm x^T\sum_{j=0}^{K-1}h_j^{i}\bm S^{j^T} \right)\right] \\&= \text{diag}\left(\sum_{k=0}^{K-1}h_k^{i}\bm S^k \mathbb{E}\left[ \bm x\bm x^T\right]\sum_{j=0}^{K-1}h_j^{i}\bm S^{j^T} \right)\nonumber\\&= \text{diag}\left(\sum_{k=0}^{K-1}h_k^{i}\bm S^k \sum_{j=0}^{K-1}h_j^{i}\bm S^{j^T} \right)
\end{align}
\subsection{Single band filtering}
In the second part of the proof we study the graph filter using the spectral decomposition of the graph:
\begin{align}
    \label{eq:GF2}
    \bm y &= \sum_{k=0}^{K-1}h_k\bm S^k \bm x \\&= \sum_{k=0}^{K-1}h_k\bm V\bm\Lambda^k\bm V^T \bm x\\&= \sum_{k=0}^{K-1}h_k\sum_{n=1}^{N}\lambda_n^k\bm v_n\bm v_n^T \bm x\\&= \sum_{n=1}^{N}\bm v_n^T \bm x\sum_{k=0}^{K-1}h_k\lambda_n^k\bm v_n.
\end{align}
Let us focus on the following polynomial:
\begin{equation}\label{eq:filterfreq}
    \tilde{h}\left(\lambda\right)=\sum_{k=0}^{K-1}h_k\lambda^k,
\end{equation}
that represents a graph filter in the frequency domain by. For $q$ distinct eigenvalues we can write a system of linear equations using the polynomial in \eqref{eq:filterfreq}:
\begin{align}
\begin{bmatrix}
\tilde{h}\left(\lambda_1\right)\\
\tilde{h}\left(\lambda_2\right)\\
\vdots\\
\tilde{h}\left(\lambda_q\right)
\end{bmatrix}=\begin{bmatrix}
1~\lambda_1~\lambda_1^2 \dots \lambda_1^{K-1}\\
1~\lambda_2~\lambda_2^2 \dots \lambda_2^{K-1}\\
\vdots\\
1~\lambda_q~\lambda_q^2 \dots \lambda_q^{K-1}\\
\end{bmatrix}\begin{bmatrix}
h_0\\
h_1\\
\vdots\\
h_{K-1}
\end{bmatrix} = \bm W \bm h
\end{align}

$\bm W$ is a Vandermonde matrix and when $K=q$ the determinant of $\bm W$ takes the form:
\begin{equation}
    \text{det}\left(\bm W\right) = \prod_{1\leq i<j\leq q} \left(\lambda_i-\lambda_j\right)
\end{equation}
Since the values $\lambda_i$ are distinct, $\bm W$ has full column rank and there exists a graph filter with unique parameters $\bm h$ that passes only the $\lambda$ eigenvalue, i.e.,
\begin{equation}\label{eq:isofilter}
    \tilde{h}\left(\lambda_i\right) = \bigg\{
\begin{matrix}
1,~\text{if}~~\lambda_i=\lambda\\
0,~\text{if}~~\lambda_i\neq \lambda
\end{matrix}
\end{equation}
Under this parametrization, equation \eqref{eq:GF2} takes the form $\bm y = \bm v_{\lambda}\bm v_{\lambda}^T\bm x$, where $\bm v_{\lambda}$ is the eigenvector corresponding to $\lambda$. 

\subsection{GNN and absolute eigenvectors}

Using the previous analysis we can design parameters $h_k$ such that:
\begin{equation}
    \sum_{k=0}^{K-1}h_k\bm S^k=\bm v_{\lambda}\bm v_{\lambda}^T
\end{equation}
and then equation \eqref{eq:expectationproof} takes the form:
\begin{align}
\mathbb{E}\left[\bm y_i\right] &= \text{diag}\left(\sum_{k=0}^{K-1}h_k^{i}\bm S^k \sum_{j=0}^{K-1}h_j^{i}\bm S^{j^T} \right)\\&=\text{diag}\left(\bm v_{\lambda}\bm v_{\lambda}^T\bm v_{\lambda}\bm v_{\lambda}^T\right) \\&= \text{diag}\left(\bm v_{\lambda}\bm v_{\lambda}^T\right) \\&= \left |\bm v_{\lambda}\right |^2
\end{align}

We can therefore design $\bm h_k\in\mathbb{R}^{m}$ for $k=0,\dots,m-1$ to compute the absolute value of $m$ eigenvectors of $\bm S$ that correspond to the top $m$ distinct eigenvalues, i.e.,
\begin{align}
    \mathbb{E}\left[\bm y_i\right] =|{\bm u}_i|^2,\quad i=1,\dots,m\\
\end{align}
We can do the same for graph $\hat{\bm S}$ and compute:
\begin{align}
    \mathbb{E}\left[\hat{\bm y}_i\right] =|\hat{\bm u}_i|^2,\quad i=1,\dots,m\\
\end{align}
Since both $\bm S,~\hat{\bm S}$ have distinct eigenvalues, we can concatenate the output of each neuron and result in layer-1 outputs as:
\begin{equation}\label{eq:proofout}
    \bm Y^{(1)} = |\bm U|,\quad \hat{\bm Y}^{(1)}=|\hat{\bm U}|
\end{equation}
As a result, the previously described GNN can a least yield the same alignment accuracy as the absolute values of the eigenvectors.
\subsection{Generalization to multiple graph pairs}
The analysis in the previoius subsections is indeed presented for a pair of graphs but can be directly extended for any set of graphs. We can generalize the Theorem \ref{theorem:abs_eig}, to read as:
Let $\{\mathcal{G}_1,\dots, \mathcal{G}_M\}$ be a set of graphs with adjacencies $\{\bm{S}_1,\dots, \bm{S}_M\}$ that have non-repeated eigenvalues. Then for any $\bm{S},~\hat{\bm{S}}\in\{\bm{S}_1,\dots, \bm{S}_M\}$, there exists a GNN $\phi\left(\bm X; \bm S, \mathcal{H} \right):\mathbb{R}^{N\times D}\to\mathbb{R}^{N\times D^{L}}$ such that:
\small{\begin{equation*}{\left\lVert~\bm S-\bm P^{\diamond}\hat{\bm S}\bm P^{\diamond^ T}\right\rVert_F^2\leq\left\lVert~\bm S-\bm P^{\ast}\hat{\bm S}\bm P^{\ast^ T}\right\rVert_F^2\leq\left\lVert~\bm S-\check{\bm P}\hat{\bm S}\check{\bm P}^T\right\rVert_F^2}
\end{equation*}}
\normalsize
with
\begin{equation*}
    \bm P^{\ast}=\argmin_{\bm P\in\mathcal{P}}~\left\lVert~\phi\left(\bm X; {\bm S}, \mathcal{H} \right)-\bm P\phi\left(\hat{\bm X}; \hat{\bm S}, \mathcal{H} \right)~\right\rVert_F^2,
\end{equation*}
where ${\bm P^\diamond},~\check{\bm P}$ are solutions to the optimization problems in (1) and (3) respectively.

\section{Implementation Details}\label{section:impSup}
In this section we discuss the implementation details of our framework.
\subsection{Assignment Optimization}
The proposed \texttt{T-GAE} learns learns a GNN encoder that can produce node representations for different graphs. Let $\phi\left({\bm X}; {\bm S}, \mathcal{H} \right)$ represent the embeddings of the nodes corresponding to the graph with adjacency $\bm S$ and $\phi\left(\hat{\bm X}; \hat{\bm S}, \mathcal{H} \right)$ represent the embeddings of the nodes corresponding to the graph with adjacency $\hat{\bm S}$. Then network alignment boils down to solving the following optimization problem:

\begin{equation}\label{eq:optimization}
\min_{\bm P\in\mathcal{P}}~\left\lVert~\phi\left(\bm X; {\bm S}, \mathcal{H} \right)-\bm P\phi\left(\hat{\bm X}; \hat{\bm S}, \mathcal{H} \right)~\right\rVert_F^2.
\end{equation}
The problem in \eqref{eq:optimization} can be optimally solved in $\mathcal{O}\left(N^3\right)$ flops by the Hungarian algorithm \citep{hungarian}. To avoid this computational burden we employ the greedy Hungarian approach that has computational complexity $\mathcal{O}\left(N^2\right)$ and usually works well in practice.

 The greedy Hungarian approach is described in Algorithm \ref{algorithmmatch}. For each row of $\phi\left({\bm X}; {\bm S}, \mathcal{H} \right),~\phi\left(\hat{\bm X}; \hat{\bm S}, \mathcal{H} \right)$, which corresponds to the node embeddings of the different graphs, we compute the pairwise Euclidean distance which is stores in the distance matrix $\bm D$. Then, at each iteration, we find the nodes with the smallest distance and remove the aligned pairs from $\bm D$. This process is repeated until all the nodes are paired up for alignment.

\begin{algorithm}[!ht]
\DontPrintSemicolon
  \KwInput{Feature matrices $\bm{X}$, $\hat{\bm{X}}$}
  \KwOutput{Assignment Matrix}
  $\bm{P}~:= \bm{0}_{N\times N}$ \tcp*{Initialize permutation matrix}
  $\bm{D}~:= $ PairwiseDistance $\left(\bm{X},\hat{\bm{X}}\right)$ \tcp*{pairwise Euclidean distance}
  $\text{rows} := $~{0,1,\dots,$N$-1} \tcp*{Corresponds to $\bm{X}$}
  $\text{cols} := $~{0,1,\dots,$N$-1} \tcp*{Corresponds to $\hat{\bm{X}}$}
  \tcc{Iterate to assign node pairs with minimum Euclidean distance}
  \For{$n$=1 to $N$}
    {
        $i,~j := \text{argmin}~(\bm{D})$\;    
        $r~:= \text{rows}~[i]$\;        
        $c~:= \text{cols}~[j]$\;       
        $\bm{P}_{rc} := 1$\;     
        Remove $r$ from rows\;      
        Remove $c$ from cols\;      
        Remove the $i$-th row from $\bm{D}$\;
        Remove the $j$-th column from $\bm{D}$\;
    }
    return $\bm{P}$

\caption{Greedy Hungarian Algorithm}
\label{algorithmmatch}
\end{algorithm}
\subsection{Datasets}
\begin{table*}[t]\centering
\setcounter{table}{1}
  \caption{Summary of Dataset statistics that are included in Section 5}
  \scalebox{0.7}{\begin{tabular}{cccccl}
    \toprule
   Task&{\text{Dataset}}&{$|\mathcal{V}|$}&{$|\mathcal{E}|$}&{$\#$ Aligned Edges}&{Network Type}\\
    \midrule
    \multirow{5}{*}{Graph Matching} &Celegans \citep{Kunegis2013KONECTTK} & 453& 2,025& 2,025&Interactome\\
    &Arenas \citep{snapnets}              & 1,133& 5,451& 5,451 &Email Communication\\
    &Cora \citep{sen2008collective}        &  2,708    &5,278&5,278    &Citation Network\\
&Douban \citep{onlineoffline}         &  3,906    &7,215&7,215 &Social Network\\
&Dblp \citep{PanWZZW16}                &  17,716   & 52,867& 52,867   &Citation Network\\
&Coauthor CS \citep{shchur2018pitfalls}&  18,333   & 81,894& 81,894 &Coauthor Network\\
  \midrule
 \multirow{4}{*}{Subraph Matching} &\thead{ACM-DBLP \citep{ACMDBLP}} & \thead{9,872\\9,916} & \thead{39,561\\44,808} & {6,352} &Citation Network\\
  
  &\thead{Douban Online-Offline \citep{onlineoffline}} & \thead{3,906\\1,118} & \thead{1,632\\3,022} & {1,118} &Social Network\\
  \bottomrule
\end{tabular}}
\label{tab:dataset}
\vspace{-0.2cm}
\end{table*}
We include statistics of the datasets used in our experiments in Table \ref{tab:dataset}. The detailed descriptions of each dataset are presented below:
\begin{itemize}
    \item Celegans \citep{Kunegis2013KONECTTK}: The vertices represent proteins and the edges their protein-protein interactions.
    \item Arenas Email \citep{snapnets}: The email communication network at the University Rovira i Virgili in Tarragona in the south of Catalonia in Spain. Nodes are users and each edge represents that at least one email was sent.  
    \item Douban \citep{onlineoffline}: Contains user-user relationship on the Chinese movie review platform. Each edge implies that two users are contacts or friends.
    \item Cora \citep{sen2008collective}: The dataset consists of 2708 scientific publications, with edges representing citation relationships between them. Cora has been one of the major benchmark datasets in many graph mining tasks. 
    \item Dblp \citep{PanWZZW16}: A citation network dataset that is extracted from DBLP, Association for Computing Machinery (ACM), Microsoft Academic Graph (MAG), and other sources. It is considered a benchmark in multiple tasks.
    \item Coauthor\_CS \citep{shchur2018pitfalls}: The coauthorship graph is generated from MAG. Nodes are the authors and they are connected with an edge if they coauthored at least one paper. 
    \item ACM$-$DBLP \citep{ACMDBLP}: The citation networks that share some common nodes. The task is to identify the publications that appear in both networks.
    \item Douban Online$-$Offline \citep{onlineoffline}: The two social networks contained in this dataset represents the online and offline events of the Douban social network. The task is to identify users that participate in both online and offline events. 
\end{itemize}
\subsection{Baselines}
\subsubsection{Graph Neural Network(GNN) based methods}
To have a fair comparison with the node embedding models, GAE is implemented using the same set of parameters as T-GAE, which can be found at Section \ref{sec::TGAEparams}. WAlign is implemented using parameters suggested by the author. We report the best results they achieved during training. 
\begin{itemize}
    \item \texttt{WAlign} \citep{jundong} fits a GNN to each of the input graphs, trains the model by reconstructing the given inputs and minimizing an approximation of Wasserstein distance between the node embeddings. We use the author's implementation from \url{https://github.com/gaoji7777/walign.git}.

    \item \texttt{GAE}, \texttt{VGAE}\citep{GAE} are self-supervised graph learning frameworks that are trained by reconstructing the graph. The encoder is a GCN\citep{GCN} and linear decoder is applied to predict the original adjacency. In VGAE, Gausian Noise is introduced to the node embeddings before passing to the decoder. We use the implementation from \url{https://github.com/DaehanKim/vgae_pytorch}. We train GAE by reconstructing the given network using the netsimile node embedding. 

\end{itemize}

\subsubsection{Graph/Node embedding techniques}
\begin{itemize}
    \item \texttt{NetSimile} \citep{netsimile} uses the structural features described earlier to match the nodes of the graphs. Since the \texttt{NetSimile} features are used as input to the \texttt{T-GAE}, they provide a measure to assess the benefit of using \texttt{T-GAE} for node embedding. It proposed 7 egonet-based features, to measure network similarity. We process these features by Algorithm \ref{algorithmmatch} to perform network alignment. The 7-dimensional Netsimile features are:
    \begin{itemize}
        \item {$d_i$ = degree of node i}
        \item {$c_i$ = number of triangles connected to node i over the number of connected triples centered on node i}
        \item {$\bar{d}_{N_{i}}$ = $\frac{1}{d_{i}}\sum_{j\in N_{i}}{d_{j}}$, average number of two-hop neighbors}
        \item {$\bar{c}_{N_{i}}$ = $\frac{1}{d_{i}}\sum_{j\in N_{i}}{c_{j}}$, average clustering coefficient}
        \item {Number of edges in node i's egonet}
        \item {Number of outgoing edges from node i's egonet}
        \item {Number of neighbors in node i's egonet}
    \end{itemize}
    The implementation is based on netrd library where we use the feature extraction function. The source code can be found at \url{https://netrd.readthedocs.io/en/latest/_modules/netrd/distance/netsimile.html}

    \item\texttt{Spectral} \citep{6778} It solves the following optimization problem:
    \begin{equation} \min_{\bm P\in\mathcal{P}}~\left\lVert~\left |\bm V\right |-\bm P\left |\hat{\bm V}\right |~\right\rVert_F^2,
    \end{equation}
    where $\bm V,~\hat{\bm V}$ are the eigenvectors corresponding to the adjacencies of the graphs that we want to match. In our initial experiments, we observed that a subset of the eigenvectors yields improved results compared to the whole set. We tried $1-10$ top eigenvectors and concluded that $4$ eigenvectors are those that yield the best results on average. Thus we solve the above problem with the top-4 eigenvectors. 

    \item \texttt{DeepWalk} \citep{deepwalk}: A node embedding approach, simulates random walks on the graph and apply skip-gram on the walks to generate node embedding. We use the implementation from \href{https://github.com/benedekrozemberczki/karateclub/blob/master/karateclub/node_embedding/neighbourhood/deepwalk.py}{Karateclub}. The algorithm is implemented with the default parameters as suggested by this repository, the number of random walks is 10 with each walk of length 80. The dimensionality of embedding is set to be 128. We run the algorithm with 1 epoch and set the learning rate to be 0.05.

    \item \texttt{Node2Vec} \citep{node2vec}: An improve version of DeepWalk, it has weights on the randomly generated random walks, to make the neighborhood preserving objective more flexible. We use the implementation from \href{https://github.com/benedekrozemberczki/karateclub/blob/master/karateclub/node_embedding/neighbourhood/node2vec.py}{Karateclub}. The default parameters are used. We simulate 10 random walks on the graph with length 80. p and q are both equal to 1. Dimensionality of embeddings is set to be 4 and we run 1 epoch with learning rate 0.05. 

    \item \texttt{GraphWave} \citep{graphwave}: The structure information of the graphs is captured by simulating heat diffusion process on them. We use the implementation from \href{https://github.com/benedekrozemberczki/karateclub/blob/master/karateclub/node_embedding/structural/graphwave.py}{Karateclub} with the default parameters: number of evaluation points is 200, step size is 0.1, heat coefficient is 1.0 and Chebyshev polynomial order is set to be 100. Note that this implementation does not work on graphs with more than 10,000 nodes, so we exclude this model on the DBLP and Coauthor\_CS dataset. 

    \item \texttt{LINE} \citep{line}: An optimization based graph embedding approach that aims to preserve local and global structures of the network by considering substructures and structural-quivariant nodes. We use the PyTorch implementation from \url{https://github.com/zxhhh97/ABot}. All parameters are set to default as the authors suggested.
\end{itemize}
\subsubsection{Optimization based graph matching algorithms}
\begin{itemize}
    \item\texttt{FINAL} \citep{onlineoffline} is an optimization approach, following an alignment consistency principle, and tries to match nodes with similar topology. In the case of graphs without attributes \texttt{FINAL} is equivalent to the popular 
    \item\texttt{Isorank} \citep{bio2} algorithm, whereas using \texttt{NetSimile} as an input to \texttt{FINAL} resulted in inferior performance and was therefore omitted. We use the code in \url{https://github.com/sizhang92/FINAL-KDD16} with $\bm H$ being the degree similarity matrix, $\alpha = 0.8,~\text{maxiter} = 30,~\text{tol} = 1e-4$ as suggested in the repository. 
    \item \texttt{ConeAlign} \citep{chen2020cone} is a graph embedding based approach. The matching is optimized in each iteration by the Wasserstein Procrustes distances between the matched embeddings calculated on a mini batch in order to preserve scalability. We use the official implementation from \url{https://github.com/GemsLab/CONE-Align} and preserved all the suggested parameters.
    \item \texttt{S-GWL} \citep{sgwl}  matches two given graphs by retrieving node correspondence from the optimal transport associated with the Gromov-Wasserstein discrepancy between the graphs. We use the implementation by the authors in \url{https://github.com/HongtengXu/s-gwl}, since the performance of S-GWL is very sensitive to the parameter gamma, as suggested by the authors, we fine-tuned this parameter over the range of $[0.001, 0.1]$ for each dataset on the cleaned graph, and use that optimal parameter for all other experiments on this dataset.
\end{itemize}

\subsection{T-GAE model details} \label{sec::TGAEparams}
As illustrated in Figure \ref{Model}, the structure of our proposed encoder consists of two MLPs and a series of GNN layers. The node features are processed by a 2-layer MLP and passed to all the GNN layers. We add skip connections between this MLP layer and all the subsequent GNN layers. The outputs of all GNN layers are concatenated and passed to another 2-layer MLP, followed by a linear decoder to generate the reconstructed graph. The model is optimized end to end by equation \ref{eq:trainingfinal}. For graph matching experiments, since we consider the general case where graphs are given without node attributes, we use the $7$ structural features proposed in \citep{netsimile}. The features include the degree of each node, the local and average clustering coefficient, and the number of edges, outgoing edges, and neighbors in each node's egonet. This input feature is applied for all GNN-based methods. As a result, the performance of \texttt{NetSimile}, vanilla \texttt{GAE} and \texttt{WAlign} provide measures to assess the benefit of using \texttt{T-GAE} for node embedding. Note that one can choose different message passing functions as $f$ and $g$ in Equation \eqref{eq:GNNrec00}, and any structure-preserving node features. Our reported results are based on \texttt{GIN} \citep{xu2018how} and \texttt{Netsimile} \citep{netsimile}. 

\section{More Baseline Results} \label{sec::more_baseline}
We present the graph matching accuracy for the baseline methods that are not permutation equavariant in Table \ref{tab:baselines_sup_gm}, and sub-graph matching accuracy on Douban Online-Offline dataset for GraphWave in Table \ref{tab:baselines_sup_sgm}, as it is not scalable on ACM/DBLP. 
\begin{table*}[h]
\centering
  \caption{Graph matching accuracy on 10 randomly perturbed samples under different levels of edge editing for VGAE, LINE and DeepWalk.}\label{tab:baselines_sup_gm}
  \small
\setlength\tabcolsep{2pt}
\scalebox{1.0}{\begin{tabular}{c|P{2cm}|P{1.7cm}|P{1.7cm}|P{1.7cm}}
    \toprule
     &{Dataset}&{VGAE}&{LINE}&{DeepWalk}\\
    \midrule
     {\multirow{2}{*}{\rotatebox[origin=c]{90}{$0\%$}}}& \cellcolor{white}Celegans & \cellcolor{red!50} ${0.3\pm0.1}$ & \cellcolor{red!50} ${1.0\pm 0.5}$ & \cellcolor{red!50} ${1.8\pm 0.6}$\\
     &\cellcolor{white}Arenas & \cellcolor{red!50}${0.1\pm0.1}$ &\cellcolor{red!50}${0.2\pm0.1}$ &\cellcolor{red!50}${0.3\pm0.2}$\\
     &\cellcolor{white}Douban &\cellcolor{red!50}${0.0\pm0.0}$ &\cellcolor{red!50}${0.0\pm0.0}$ &\cellcolor{red!50}${0.1\pm0.0}$ \\
     &\cellcolor{white}Cora &\cellcolor{red!50}${0.1\pm0.0}$ &\cellcolor{red!50}${0.0\pm0.0}$ &\cellcolor{red!50}${0.1\pm0.0}$\\
     \midrule
     {\multirow{2}{*}{\rotatebox[origin=c]{90}{$1\%$}}}& \cellcolor{white}Celegans & \cellcolor{red!50} ${0.3\pm0.1}$ & \cellcolor{red!50} ${1.0\pm 0.4}$ & \cellcolor{red!50} ${1.2\pm 0.5}$\\
     &\cellcolor{white}Arenas & \cellcolor{red!50}${0.1\pm0.1}$ &\cellcolor{red!50}${0.1\pm0.1}$ &\cellcolor{red!50}${0.3\pm0.1}$\\
     &\cellcolor{white}Douban &\cellcolor{red!50}${0.0\pm0.0}$ &\cellcolor{red!50}${0.0\pm0.0}$ &\cellcolor{red!50}${0.1\pm0.0}$ \\
     &\cellcolor{white}Cora &\cellcolor{red!50}${0.1\pm0.1}$ &\cellcolor{red!50}${0.1\pm0.0}$ &\cellcolor{red!50}${0.2\pm0.1}$\\
     \midrule
     {\multirow{2}{*}{\rotatebox[origin=c]{90}{$5\%$}}}& \cellcolor{white}Celegans & \cellcolor{red!50} ${0.6\pm0.3}$ & \cellcolor{red!50} ${0.9\pm 0.3}$ & \cellcolor{red!50} ${1.0\pm 0.3}$\\
     &\cellcolor{white}Arenas & \cellcolor{red!50}${0.2\pm0.1}$ &\cellcolor{red!50}${0.2\pm0.2}$ &\cellcolor{red!50}${0.2\pm0.1}$\\
     &\cellcolor{white}Douban &\cellcolor{red!50}${0.0\pm0.0}$ &\cellcolor{red!50}${0.0\pm0.0}$ &\cellcolor{red!50}${0.0\pm0.0}$ \\
     &\cellcolor{white}Cora &\cellcolor{red!50}${0.1\pm0.0}$ &\cellcolor{red!50}${0.1\pm0.0}$ &\cellcolor{red!50}${0.1\pm0.0}$\\
    \bottomrule
\end{tabular}}
\end{table*}
\begin{table*}[h]
\centering
  \caption{Sub-graph matching performance for GraphWave on Douban Online-Offline}\label{tab:baselines_sup_sgm}
  \small
\setlength\tabcolsep{2pt}
\scalebox{1.0}{\begin{tabular}{c|P{2cm}|P{1.7cm}}
    \toprule
     &{Hit rate}&{GraphWave}\\
    \midrule
     &\cellcolor{white}Hit@1 & ${0.09}$\\
     &\cellcolor{white}Hit@5 & ${0.36}$\\
     &\cellcolor{white}Hit@10 &${0.81}$ \\
     &\cellcolor{white}Hit@50 &${4.74}$\\
     &\cellcolor{white}Hit@100 &${9.12}$\\
    \bottomrule
\end{tabular}}
\end{table*}

\section{Degree Perturbation Model Results}\label{section:degreeModel}

\begin{table*}[h]
\centering
  \caption{Graph matching accuracy on 10 randomly perturbed samples under different levels of edge removal on Degree model. The proposed T-GAE is trained on the clean Celegans, Arena, Douban, and Cora networks, and tested on noisy versions of them and the larger Dblp, and Coauthor CS. Accuracy above 80\% is highlighted in green, 60\% to 80\% accuracy is in yellow, and performance below 60\% is in red.}\label{tab:transferdegree}
  \small
\setlength\tabcolsep{2pt}
\scalebox{0.7}{\begin{tabular}{c|P{2.7cm}|P{1.7cm}|P{1.7cm}|P{1.7cm}|P{1.7cm}|P{1.7cm}|P{1.7cm}|P{1.7cm}|P{1.7cm}|P{1.7cm}}
    \toprule
      &\multirow{2}{*}{Dataset $\backslash$ Algorithm}&
      \multicolumn{3}{c|}{\textit{\textbf{Feature Engineering based}}} &
      \multicolumn{3}{c|}{\textit{\textbf{Optimization based}}} &
      \multicolumn{3}{c}{\textit{\textbf{GNN based}}} \\
   &&{Spectral}&{Netsimile}&{GraphWave}&{FINAL}&{S-GWL}&{ConeAlign}&{WAlign}&{GAE}&\textbf{T-GAE}\\
    \midrule
      {\multirow{6}{*}{\rotatebox[origin=c]{90}{$0\%$ perturbation}}}& \cellcolor{white}Celegans & \cellcolor{green!50} ${87.8\pm1.5}$ & \cellcolor{yellow!50} ${72.7\pm 0.9}$ & \cellcolor{yellow!50} ${65.3\pm1.7}$ & \cellcolor{green!50} ${92.2\pm 1.2}$& \cellcolor{green!50}$\mathbf{93.0\pm1.5}$ & \cellcolor{yellow!50} ${66.6\pm1.2}$ &\cellcolor{green!50}${88.4\pm1.6}$ &\cellcolor{green!50}${90.9\pm2.6}$ &\cellcolor{green!50}${91.0\pm1.1}$\\
     
     &\cellcolor{white}Arenas & \cellcolor{green!50}${97.7\pm0.4}$ &\cellcolor{green!50}${94.7\pm0.3}$  &\cellcolor{green!50}${81.7\pm0.7}$ &\cellcolor{green!50}${97.5\pm0.3}$&\cellcolor{green!50}${97.5\pm0.3}$  &\cellcolor{green!50}${87.8\pm0.6}$&\cellcolor{green!50}${97.4\pm0.5}$ &\cellcolor{green!50}${97.6\pm0.4}$ &\cellcolor{green!50}$\mathbf{97.8\pm0.4}$\\
     
     &\cellcolor{white}Douban &\cellcolor{green!50}${89.9\pm0.4}$ &\cellcolor{yellow!50}${46.4\pm0.4}$  &\cellcolor{red!50}${17.5\pm0.2}$ &\cellcolor{green!50}${89.9\pm0.3}$&\cellcolor{green!50}$\mathbf{90.1\pm0.3}$ &\cellcolor{yellow!50}${68.1\pm0.4}$ &\cellcolor{green!50}${90.0\pm0.4}$ &\cellcolor{green!50}${89.5\pm0.4}$ &\cellcolor{green!50}$\mathbf{90.1\pm0.3}$\\
     
     &\cellcolor{white}Cora &\cellcolor{green!50}${85.0\pm0.4}$ &\cellcolor{yellow!50}${73.7\pm0.4}$ &\cellcolor{red!50}${8.3\pm0.4}$&\cellcolor{green!50}${87.5\pm0.7}$ &\cellcolor{green!50}${87.3\pm0.7}$ &\cellcolor{red!50}${38.5\pm0.7}$ &\cellcolor{green!50}${87.2\pm0.4}$ &\cellcolor{green!50}${87.1\pm0.8}$ &\cellcolor{green!50}$\mathbf{87.5\pm0.4}$\\
     
     &\cellcolor{white}Dblp &\cellcolor{green!50}${84.5\pm0.1}$ &\cellcolor{yellow!50}${63.7\pm0.2}$  &\cellcolor{red!50}doesn't scale &\cellcolor{green!50}$\mathbf{85.6\pm0.2}$&\cellcolor{red!50}> 48 hours &\cellcolor{red!50}${44.3\pm0.6}$ &\cellcolor{green!50}${85.6\pm0.2}$ &\cellcolor{green!50}${85.2\pm0.3}$  &\cellcolor{green!50}$\mathbf{85.6\pm0.2}$\\
     
     &\cellcolor{white}Coauthor CS &\cellcolor{green!50}${97.5\pm0.1}$ &\cellcolor{green!50}${90.9\pm0.1}$ &\cellcolor{red!50}doesn't scale &\cellcolor{green!50}$\mathbf{97.6\pm0.1}$ &\cellcolor{red!50}> 48 hours &\cellcolor{yellow!50}${75.8\pm0.5}$ &\cellcolor{green!50}${97.5\pm0.2}$ &\cellcolor{green!50}${97.6\pm0.3}$ &\cellcolor{green!50}$\mathbf{97.6\pm0.1}$\\
     \midrule
     {\multirow{6}{*}{\rotatebox[origin=c]{90}{$1\%$ perturbation}}}& \cellcolor{white}Celegans & \cellcolor{red!50} ${21.8\pm16.9}$ & \cellcolor{yellow!50} ${63.1\pm 1.4}$  & \cellcolor{red!50} ${4.9\pm1.1}$ & \cellcolor{yellow!50} ${62.5\pm 4.1}$& \cellcolor{yellow!50}$\mathbf{73.1\pm11.7}$ & \cellcolor{yellow!50} ${61.4\pm3.2}$ &\cellcolor{yellow!50}${70.7\pm4.4}$ &\cellcolor{red!50}${7.8\pm2.5}$ &\cellcolor{yellow!50}${63.4\pm6.1}$\\
     
     &\cellcolor{white}Arenas & \cellcolor{yellow!50}${66.0\pm21.6}$ &\cellcolor{green!50}${90.9\pm0.7}$ &\cellcolor{red!50}${5.8\pm2.6}$ &\cellcolor{red!50}${56.7\pm2.7}$&\cellcolor{green!50}${92.3\pm1.2}$ &\cellcolor{green!50}${86.1\pm0.6}$ &\cellcolor{green!50}${96.0\pm0.5}$ &\cellcolor{red!50}${0.9\pm0.5}$ &\cellcolor{green!50}$\mathbf{96.7\pm0.6}$\\
     
     &\cellcolor{white}Douban &\cellcolor{red!50}${9.8\pm13.5}$ &\cellcolor{red!50}${37.6\pm0.6}$ &\cellcolor{red!50}${0.6\pm0.3}$ &\cellcolor{red!50}${35.0\pm1.1}$ &\cellcolor{yellow!50}${69.9\pm1.2}$ &\cellcolor{red!50}${59.6\pm2.2}$ &\cellcolor{green!50}${81.3\pm1.9}$ &\cellcolor{red!50}${0.4\pm0.0}$ &\cellcolor{green!50}$\mathbf{83.7\pm2.2}$\\
     
     &\cellcolor{white}Cora &\cellcolor{red!50}${25.3\pm13.4}$ &\cellcolor{yellow!50}${64.4\pm1.2}$ &\cellcolor{red!50}${0.1\pm0.0}$ &\cellcolor{red!50}${32.0\pm2.0}$&\cellcolor{red!50}${31.1\pm4.4}$ &\cellcolor{red!50}${30.6\pm2.7}$ &\cellcolor{yellow!50}${74.6\pm0.7}$ &\cellcolor{red!50}${28.2\pm0.3}$ &\cellcolor{green!50}$\mathbf{81.0\pm2.5}$\\
     
     &\cellcolor{white}Dblp &\cellcolor{red!50}${3.4\pm0.8}$ &\cellcolor{red!50}${52.2\pm0.5}$ &\cellcolor{red!50}doesn't scale &\cellcolor{red!50}${24.5\pm0.6}$ &\cellcolor{red!50}> 48 hours &\cellcolor{red!50}${21.1\pm1.9}$ &\cellcolor{yellow!50}${67.4\pm1.1}$ &\cellcolor{red!50}${10.5\pm0.6}$  &\cellcolor{yellow!50}$\mathbf{77.1\pm0.6}$\\
     
     &\cellcolor{white}Coauthor CS &\cellcolor{red!50}${8.6\pm4.6}$ &\cellcolor{yellow!50}${76.8\pm0.6}$   &\cellcolor{red!50}doesn't scale &\cellcolor{red!50}${19.1\pm0.5}$ &\cellcolor{red!50}> 48 hours &\cellcolor{yellow!50}${64.0\pm2.0}$ &\cellcolor{green!50}${88.9\pm0.8}$ &\cellcolor{red!50}${3.8\pm0.1}$ &\cellcolor{green!50}$\mathbf{94.0\pm0.4}$\\
    \bottomrule
\end{tabular}}
\end{table*}

 \noindent \textbf{Degree Model:} {In this model we only remove edges. Edges with higher degrees are more likely to be removed to preserve the structure of the graph. Specifically, the probability of removing edge $(i,j)$ is set to  $\frac{s_{ij}d_{i}d_{j}}{\sum_{ij}{s_{ij}d_{i}d_{j}}}$, where $d_{i}$ is the degree of node $v_i$ and $s_{i,j}$ is the $(i,j)$ element of the graph adjacency.
 
We test the performance of \texttt{T-GAE} for large-scale network alignment on the degree perturbation model, as described in Section \ref{section:pert}. We adopt the same setting as in Section \ref{section:pert} to train the \texttt{T-GAE} according to \eqref{eq:family} on small-size networks, i.e., Celegans, Arena, Douban, and Cora, and conduct transfer learning experiments on the larger graphs, i.e., Dblp, and Coauthor CS. The trained T-GAE is used to generate node embedding for the graphs, and Algorithm \ref{algorithmmatch} computes the assignment matrix. The results presented in Table \ref{tab:transferdegree} are based on the average and standard deviation of the matching accuracy over 10 randomly generated perturbed samples. 

We observe that the benefit of processing the \texttt{NetSimile} embeddings with GNNs is still significant in this perturbation model as we observe up to $46\%$ performance increase at the presence of perturbation. When testing on perturbed graphs at $1\%$ level of edge removal, our proposed \texttt{T-GAE} consistently outperforms all the competing baselines, while being robust and efficient when performing network alignment under different perturbation models.
Especially, on large-scale networks, \texttt{T-GAE} is able to achieve very high levels of matching accuracy for both Dblp and Coauthor CS, for $p=0\%,~1\%$. 


The benefit of our proposed \texttt{T-GAE} framework in improving the expressiveness of GNN still stands out if we compare the accuracy with WAlign and GAE. It also consistently achieves the best result among all baseline methods that are salable.

\section{Efficiency analysis}\label{sec::efficiency}
\subsection{Running time comparison}
\textbf{T-GAE is a scalable and efficient approach for network alignment.} We analyze the efficiency of the proposed graph matching framework by comparing its running time with the competing algorithms. T-GAE achieves at most $\times$2000 less running time, on graph matching tasks, compared to the optimization-based methods, as shown in Table \ref{runtime_graph_matching}. Compared to the existing GNN based approaches, T-GAE consistently achieves the shortest training time and inference time, this is because we replace some message passing layers by the local MLP which serves as an attention function on the output of all GNN layers, such models are empirically proved to be more efficient, at the same time, prompting the expressiveness of GNN layers, to generate node embedding for graph matching. It should be noted that T-GAE is also the only approach that is transferable, which means it does not need to re-train on every pair of new graphs. T-GAE greatly improves the scalability of optimization based methods, as well as the effectiveness of the existing GNN frameworks.
\begin{figure}[H]
        \centering
         \includegraphics[height=10cm, width=\linewidth]{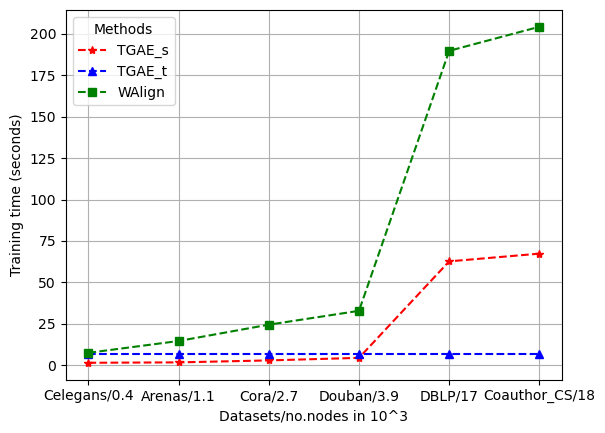}
    \caption{Training time comparison (20 epoches) between T-GAE and WAlign for graph-matching. $\text{TGAE}\_{\text{s}}$ is the specific setting where we train the encoder GNN according to Equation (7), whereas $\text{TGAE}\_{\text{t}}$ means training according to Equation (8) on a family of graphs. (Celegans, Arenas, Cora, Douban)}
\label{trainingtimepic}
\end{figure}

The proposed training objective \eqref{eq:trainingfinal} scales well from networks with 400 nodes \citep{Kunegis2013KONECTTK} to denser networks with $\times$50 nodes \citep{shchur2018pitfalls}, with minor running time increase, compared to other GNN-based frameworks(WAlign), as shown in Figure \ref{trainingtimepic}. 
\begin{table*}
\centering
  \caption{Runtime(inference+matching) in seconds for the competing algorithms on graph matching tasks.}\label{runtime_graph_matching}
  \small
\setlength\tabcolsep{2pt}
\begin{tabular}{p{3.2 cm}|p{1.7cm}|p{1.7cm}|p{1.7cm}|p{1.7cm}|p{1.7cm}|p{1.7cm}}
    \toprule
     {Algorithm}&{Celegans}&{Arenas}&{Cora}&{Douban}&{Dblp}&{Coauthor CS}\\
     \midrule
     Spectrul & ${5.712}$ &${2.819}$ &${54.770}$ &${60.298}$  &> 48 hours &> 48 hours\\
     Netsimile & ${1.212}$ & ${1.560}$  & ${1.616}$ & ${4.400}$ & ${195.542}$ & ${225.546}$\\
     GraphWave & ${8.308}$ &${32.994}$ &${131.230}$ &${281.629}$  &${5470.724}$ &${6291.368}$\\
     \midrule
     FINAL(Matlab) &${0.030}$ &${0.081}$ &${0.498}$ &${1.007}$  &${86.788}$ &${118.065}$\\
     S-GWL &${27.844}$ &${37.443}$ &${311.201}$ &${3394.522}$  &> 48 hours &> 48 hours\\
     ConeAlign & ${1.333}$ &${3.500}$ &${13.799}$ &${31.955}$  &${887.090}$ &${1099.145}$\\
     \midrule
     WAlign & ${0.078}$ & ${0.205}$ &${0.766}$ &${1.800}$  &${169.694}$ &${189.032}$\\
     GAE  & ${0.074}$ & ${0.212}$ & ${0.757}$ & ${1.749}$  &${164.410}$ &${184.062}$\\
     T-GAE(ours) & $\mathbf{0.068}$ & $\mathbf{0.201}$ & $\mathbf{0.742}$ & $\mathbf{1.734}$  &$\mathbf{163.836}$ &$\mathbf{183.289}$\\
    \bottomrule
\end{tabular}
\end{table*}
\subsection{Matching algorithms comparison}

In this subsection, we evaluate the accuracy and matching time of different matching algorithms, to demonstrate how matching algorithms of different time complexity influence the performance of the proposed T-GAE framework. To guarantee fairness of comparison, we use an untrained T-GAE encoder to encode the graphs, and use (1) approximated NN algorithm introduced in Section \ref{sec::alignment_complexity} of time complexity $\mathcal{O}(NlogN)$. (2) greedy Hungarian algorithm as applied in Section \ref{sec::main_exp} and \ref{sec::subgraph} of time complexity $\mathcal{O}(N^2)$. (3) exact Hungarian algorithm \citep{hungarian} of time complexity $\mathcal{O}(N^3)$, where $N$ is the number of nodes in the graph. We report average accuracy and running time on matching 10 randomly generated samples. 
\newcommand{\centered}[1]{\begin{tabular}{l} #1 \end{tabular}}
\definecolor{deepgreen}{rgb}{0.0, 0.5, 0.0}
\begin{table}[ht]
\caption{ 
Graph matching performance and matching time on Celegans and Arenas using untrained T-GAE encoder. We report results of approximated NN matching algorithm, greedy Hungarian algorithm, and exact Hungarian algorithm. We highlight the performance gain of exact Hungarian over the approximated NN and Greedy Hungarian.
}
\label{tab::matching_alg}
\begin{center}
\resizebox{\linewidth}{!}{
{
\setlength\tabcolsep{3.5pt}
\begin{tabular}{rl|cc|cc|cc|cc|cc|cc|}

\toprule

\multirow{2}{*}[-1pt]{\#} & \multirow{2}{*}[-1pt]{\makecell[l]{Dataset/Perturbation}} &
\multicolumn{6}{c|}{Celegans} & \multicolumn{6}{c|}{Arenas}
\\ 
&&
\multicolumn{2}{c|}{0} & \multicolumn{2}{c|}{0.01} & \multicolumn{2}{c|}{0.05} & \multicolumn{2}{c|}{0} & \multicolumn{2}{c|}{0.01} & \multicolumn{2}{c|}{0.05}\\
 && \centered{Acc} & Time & Acc & Time & Acc & Time & Acc & Time & Acc & Time & Acc & Time 
\\
\midrule
&&\multicolumn{12}{c|}{\centered{\textit{\textbf{$\mathcal{O}(NlogN)$}}}}\\
1 & Approximated NN
& $89.8\pm1.0$ & 0.004 
& $0.8\pm0.3$ & 0.004
& $0.7\pm0.2$ & 0.004
& $98.0\pm2.6$ & 0.009
& $0.1\pm0.1$ & 0.009
& $0.0\pm0.0$ & 0.009
\\ 
\midrule
&&\multicolumn{12}{c|}{\centered{\textit{\textbf{$\mathcal{O}(N^2)$}}}}\\
2 & Greedy Hungarian
& $88.4\pm0.9$ & 0.068 
& $80.3\pm0.3$ & 0.068
& $58.2\pm3.5$ & 0.067
& $97.6\pm0.4$ & 0.173 
& $93.1\pm0.5$ & 0.176
& $60.2\pm5.2$ & 0.174
\\
\midrule
&&\multicolumn{12}{c|}{\centered{\textit{\textbf{$\mathcal{O}(N^3)$}}}}\\
3 & Exact Hungarian
& $88.5\pm0.8$ & 0.225 
& $84.0\pm0.2$ & 24.778
& $64.7\pm1.8$ & 73.674
& $97.6\pm0.4$ & 0.411 
& $93.8\pm0.4$ & 562.221
& $70.0\pm2.0$ & 1624.713
\\
\midrule
&&\multicolumn{12}{c|}{\centered{\textit{\textbf{$\mathcal{O}(N^3)$}}}}\\
4 & Acc Gain in \% (Exact vs Approx/Greedy)
& \multicolumn{2}{c|}{\color{red}{-1.3} / \color{deepgreen}{+0.1} } & \multicolumn{2}{c|}{\color{deepgreen}{+83.2 / +3.7}} & \multicolumn{2}{c|}{\color{deepgreen}{+64.0 / +6.5}} & \multicolumn{2}{c|}{\color{red}{-0.4} / 0} & \multicolumn{2}{c|}{\color{deepgreen} +93.7 / +0.7} & \multicolumn{2}{c|}{\color{deepgreen} +70.0 / +9.8}\\
\bottomrule
\end{tabular}
}}
\end{center}
\end{table}

\textbf{The performance of T-GAE to match small graphs can be further improved by adopting the exact Hungarian algorithm.} Comparing Exact Hungarian and Greedy Hungarian, we observe that when there is no perturbation involved, greedy Hungarian and exact Hungarian achieves comparable performance. However, the exact Hungarian algorithm outperforms the greedy version in occurrence of perturbations, and the performance gap increases as we introduce more topology noise. Specifically, it offers a 6.5\% and 9.8\% matching accuracy increase on Celegans and Arenas, respectively, on an untrained T-GAE encoder. However, it can take more than $\times$9000 longer than the greedy algorithm, on Arenas 5\% perturbation, for example. This implies that whenever applicable, the exact Hungarian algorithm should be applied to further improve the performance of T-GAE to match small graphs, especially when the noise level is high, but there is a trade-off between matching accuracy and efficiency. 

\textbf{The efficiency of T-GAE to match permuted graphs can be enhanced by applying the approximated NN algorithm. } We observe that the approximated NN algorithm we introduced in Section \ref{sec::alignment_complexity} effectively match the aligned nodes of permuted graphs, and on Arenas dataset of 1,133 nodes, it saves 95\% matching time compared to the greedy Hungarian algorithm. However, this algorithm fails to match perturbed graphs. This is because the 1-dimensional feature is not expressive enough to to catch the topological noise. Our experiments prove that this efficient NN algorithm should be applied when we match very large scale networks with their permuted versions.

Overall, we divide graph matching using T-GAE in three scenarios: (1) Matching small graphs, exact Hungarian algorithm should be applied. (2) Matching large scale networks without perturbation, the approximation NN algorithm should be deployed to enhance efficiency. (3) The greedy Hungarian algorithm provides a good trade-off between time and efficiency for the general form of graph matching.  

\end{document}